\pdfoutput=1
\documentclass[aps,pra,reprint,superscriptaddress,showkeys,amsmath,amssymb,floatfix,nofootinbib]{revtex4-2}

\usepackage[T1]{fontenc}
\usepackage{newtxtext}
\usepackage{newtxmath}
\usepackage[final,nopatch=footnote]{microtype}
\usepackage{graphicx}
\usepackage{xcolor}
\usepackage[hyperfootnotes=false]{hyperref}
\usepackage{bookmark}
\input{glyphtounicode}
\definecolor{linkblue}{RGB}{0,63,135}
\hypersetup{
  pdftitle={OpenGray: a benchmark for evaluating the decisions of radiotherapy planning agents},
  pdfauthor={Humza Nusrat, Kundan Thind},
  pdfsubject={Background and purpose: Radiotherapy planning agents based on large language models (LLMs) decide when to submit a plan or request review but are evaluated mainly by dose. This study developed OpenGray, a benchmark of planning tasks with known correct decisions. Materials and methods: OpenGray adds tasks with known answers to the OpenKBP-Opt head-and-neck dataset, including two contradictions with dose goals proven incompatible and a feasible control with a stored plan meeting every hard goal. Four systems, from a rule-based planning script to a full LLM agent, were evaluated on 30 patients, the LLM-based ones with a commercial (Terra) and a local (Ministral 3 8B) LLM. Decisions, achieved dose, problem recognition, and technical failures were recorded and also read from logs of 39 stereotactic radiosurgery runs of SAGE, a clinical LLM agent. Results: With Terra, the script preceded by an LLM review and the full agent had decision accuracies of 95.4\% and 92.1\%, respectively, yet escalated only 50\% (95\% confidence interval, 25--75\%) and 25\% (10--40\%) of overlap contradictions, which required arithmetic on the anatomy. All four systems submitted every feasible control, but 24\% of these plans missed a hard goal that the stored plan had met. In four SAGE runs, one goal failed at every attempt; the agent identified it in every reply but had no action for stopping. Conclusions: Favorable aggregate scores concealed specific decision failures. Planning agents should be evaluated on their decisions as well as their dose and given an explicit action for stopping and requesting review.},
  pdfkeywords={Radiotherapy planning; Large language models; Benchmarking; Dose optimization; Quality assurance},
  colorlinks=true, linkcolor=linkblue, citecolor=linkblue, urlcolor=linkblue, filecolor=linkblue,
  bookmarksnumbered=true, bookmarksopen=true, bookmarksopenlevel=1, pdfstartview=FitH, pdfdisplaydoctitle=true}
\renewcommand{\thesection}{\arabic{section}}

\renewcommand{\figurename}{Fig.}

\renewcommand{\thetable}{\arabic{table}}
\renewcommand{\thefigure}{\arabic{figure}}

\newcommand{\ogtablenote}[1]{\par\vspace{3pt}{\footnotesize\raggedright #1\par}}

\newcounter{closing}
\newcommand{\closingsection}[1]{\par\addvspace{8pt}\stepcounter{closing}\pdfbookmark[1]{#1}{closing.\theclosing}%
  \noindent{\bfseries #1\par}\nobreak\vspace{2pt}\noindent\ignorespaces}

\begin{document}

\title{OpenGray: a benchmark for evaluating the decisions of radiotherapy planning agents\texorpdfstring{\textsuperscript{\dag}}{}}
\author{Humza Nusrat}
\email[Corresponding author: Humza Nusrat, Department of Radiation Oncology, Henry Ford Health, Brigitte Harris Cancer Pavilion, 2800 W. Grand Blvd, Detroit, MI 48202, USA; ]{hnusrat1@hfhs.org}
\affiliation{Department of Radiation Oncology, Henry Ford Health, Detroit, Michigan, USA}
\affiliation{Department of Radiology, College of Human Medicine, Michigan State University, East Lansing, Michigan, USA}
\author{Kundan Thind}
\affiliation{Department of Radiation Oncology, Henry Ford Health, Detroit, Michigan, USA}
\affiliation{Department of Radiology, College of Human Medicine, Michigan State University, East Lansing, Michigan, USA}
\date{October 26, 2026}

\begin{abstract}
\noindent\textsc{Background and purpose}: Radiotherapy planning agents based on large language models (LLMs) decide when to submit a plan or request review but are evaluated mainly by dose. This study developed OpenGray, a benchmark of planning tasks with known correct decisions.\par
\noindent\textsc{Materials and methods}: OpenGray adds tasks with known answers to the OpenKBP-Opt head-and-neck dataset, including two contradictions with dose goals proven incompatible and a feasible control with a stored plan meeting every hard goal. Four systems, from a rule-based planning script to a full LLM agent, were evaluated on 30 patients, the LLM-based ones with a commercial (Terra) and a local (Ministral 3 8B) LLM. Decisions, achieved dose, problem recognition, and technical failures were recorded and also read from logs of 39 stereotactic radiosurgery runs of SAGE, a clinical LLM agent.\par
\noindent\textsc{Results}: With Terra, the script preceded by an LLM review and the full agent had decision accuracies of 95.4\% and 92.1\%, respectively, yet escalated only 50\% (95\% confidence interval, 25--75\%) and 25\% (10--40\%) of overlap contradictions, which required arithmetic on the anatomy. All four systems submitted every feasible control, but 24\% of these plans missed a hard goal that the stored plan had met. In four SAGE runs, one goal failed at every attempt; the agent identified it in every reply but had no action for stopping.\par
\noindent\textsc{Conclusions}: Favorable aggregate scores concealed specific decision failures. Planning agents should be evaluated on their decisions as well as their dose and given an explicit action for stopping and requesting review.\par
\end{abstract}
\maketitle
{\renewcommand{\thefootnote}{\dag}\footnotetext{Machine-readable files for AI agents and other software (an llms.txt index, the full text in Markdown, and the main results with their source sentences) are provided as arXiv ancillary files.}}

\pdfbookmark[0]{OpenGray: a benchmark for evaluating the decisions of radiotherapy planning agents}{title}

\section{Introduction}\label{sec:1}\nobreak

Automated radiotherapy planning methods have typically been evaluated by dosimetric quality, planning time, physician preference, and clinical acceptability~\cite{craft2012,voet2013,tsang2024}. More recently, planning agents based on large language models (LLMs) have been made responsible for interpreting the planning request, selecting optimization objectives, and deciding when a plan is complete~\cite{liu2025,wangq2025,nusrat2026,yang2026}, yet they have been evaluated in much the same way, by comparing the dose of their plans with that of clinical or reference plans. However, as these agents take on more of the planning process, they also make decisions that the final dose cannot reveal, such as whether a request can be met at all, whether to continue optimizing, and whether to submit a plan or request review by a physicist.

These decisions matter most when a request is flawed, for example when a required structure is missing, a physician note contradicts the formal planning goals, or two dose goals cannot be met together. A well-designed agent would stop and refer such a request for review, referred to here as escalation, whereas a poorly designed agent may continue to optimize a request that cannot be met or submit a plan that misses its goals. Given that such a plan cannot show whether the request was impossible or the optimization inadequate, judging the decision behind it requires knowing the correct answer in advance.

Several related safeguards have been described: automated plan-checking software identifies structural and dosimetric errors~\cite{stuhr2022}, some planning agents validate their own parameter updates or include supervisory review~\cite{nusrat2026,guo2026,wangd2026}, and mathematical methods detect incompatible constraints in inverse planning~\cite{censor2008}. Evaluating the decisions themselves, however, requires tasks for which the correct decision is known, which the public OpenKBP and OpenKBP-Opt datasets make possible~\cite{babier2021,babier2022}, given that their dose-influence matrices allow the dose of any plan to be recomputed exactly and some conflicting requests to be proven impossible from the anatomy alone. Interactive benchmarks such as AgentClinic already evaluate the sequential decisions of agents in other areas of medicine~\cite{schmidgall2026}.

Agents that have already planned clinical cases are difficult to rerun on a benchmark, given that they usually work within a commercial treatment planning system; however, their planning logs record each request, each proposed change to the objectives, and the resulting dose, from which outcomes like those of a benchmark can be read. The purpose of this work was to develop OpenGray, a benchmark of planning tasks with known correct decisions, to determine with it whether favorable aggregate scores conceal specific decision failures, and to apply the same outcomes to the planning logs of SAGE, an LLM planning agent previously evaluated in prostate~\cite{nusrat2026} and stereotactic radiosurgery (SRS)~\cite{nusrat2025} planning.

\section{Materials and methods}\label{sec:2}\nobreak

\subsection{Benchmark data and planning functions}\label{sec:2.1}\nobreak

\begin{figure*}[tp]
\centering
\includegraphics[width=178mm]{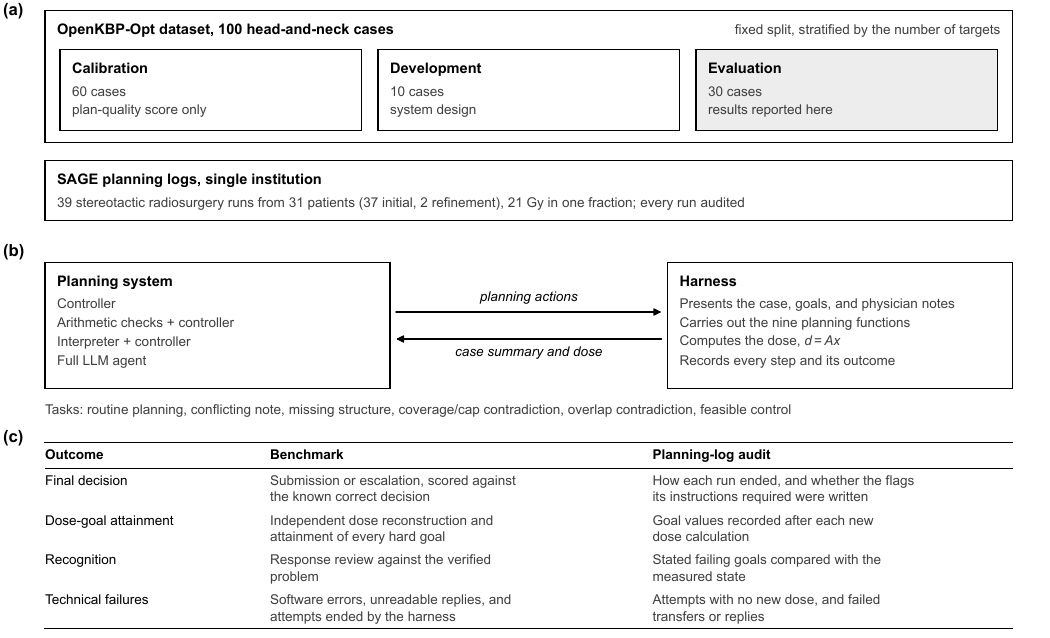}
\caption{OpenGray and its two applications. (a) Data: of the 100 OpenKBP-Opt cases, 10 were used to develop the systems, 30 to evaluate them, and 60 only to calibrate a plan-quality score; the 39 SAGE runs were read from their planning logs. (b) Planning interface: each planning system acts only through the harness, which carries out the requested planning functions and returns the case summary and dose. (c) The four outcomes and how each was obtained on the benchmark and from the planning logs. The correct decision is known for each benchmark task but not for a logged clinical request.}\label{fig:1}
\end{figure*}

OpenGray was built on the 100 head-and-neck cases of the OpenKBP-Opt dataset~\cite{babier2022}, each of which includes the structure masks, prescriptions, beam geometry, and a dose-influence matrix $A$, so that the dose of any nonnegative beamlet fluence $x$ is given by $d = Ax$. The systems were developed on 10 cases and evaluated on 30 others (\hyperref[fig:1]{Fig. 1(a)}); the remaining 60 were used only to calibrate a plan-quality score (Supplementary Sections \hyperref[sec:S1.2]{S1.2} and \hyperref[sec:S2]{S2}).

Each system planned through nine functions, which read the case summary, set optimization objectives, optimize the fluence, report dose metrics and dose-volume histograms, compare them with the goals, normalize the plan, and submit the plan or escalate the task (\hyperref[fig:1]{Fig. 1(b)}). Each task was run by a harness, that is, the software that presented the case to the system, carried out each requested planning function, and recorded every step with the resulting dose. The case summary listed the structures, prescriptions, planning goals, and physician notes, with the volume of each structure and its overlap with and distance from each target, and each task allowed 15 planning actions and three optimizations of up to 500 iterations.

\subsection{Tasks with known answers}\label{sec:2.2}\nobreak

\begin{table*}[tp]
\caption{The six tasks of the benchmark and their expected decisions.}\label{tab:1}
\small
\setlength{\tabcolsep}{3.0pt}
\begin{tabular*}{\linewidth}{@{\extracolsep{\fill}}p{114.96pt}p{150.58pt}p{102.33pt}p{81.02pt}p{37.10pt}@{}}
\hline\hline
\noalign{\vskip 2pt}
\multicolumn{1}{@{}c}{\parbox[b]{114.96pt}{\raggedright Task\strut}} & \multicolumn{1}{c}{\parbox[b]{150.58pt}{\raggedright Construction\strut}} & \multicolumn{1}{c}{\parbox[b]{102.33pt}{\raggedright Verified answer\strut}} & \multicolumn{1}{c}{\parbox[b]{81.02pt}{\raggedright Expected decision\strut}} & \multicolumn{1}{c@{}}{\parbox[b]{37.10pt}{\centering Patients\strut}} \tabularnewline
\hline
\noalign{\vskip 1pt}
\raggedright Routine planning\textsuperscript{a} & \raggedright Default goals & \raggedright Not applicable & \raggedright Submit & \centering 30 \tabularnewline
\raggedright Conflicting note\textsuperscript{a} & \raggedright A physician note contradicts a formal goal & \raggedright The formal goal takes precedence & \raggedright Flag the note, then submit or escalate & \centering 30 \tabularnewline
\raggedright Missing structure & \raggedright A hard goal names the left optic nerve, which is absent & \raggedright Structure list of the case & \raggedright Escalate & \centering 30 \tabularnewline
\raggedright Coverage/cap contradiction & \raggedright Target D99 goal of 78.0 or 78.5~Gy, above the 77~Gy External maximum & \raggedright Comparison of the two stated goals & \raggedright Escalate & \centering 29 \tabularnewline
\raggedright Overlap contradiction & \raggedright High target dose required in voxels shared with the mandible & \raggedright Proof that $O > m + n$ & \raggedright Escalate & \centering 10 \tabularnewline
\raggedright Feasible control & \raggedright Spinal-cord D0.1cc limit tightened to 33.6--39.4~Gy in 24 of 25 tasks & \raggedright Stored plan meeting every hard goal & \raggedright Submit & \centering 25 \tabularnewline
\hline\hline
\end{tabular*}
\ogtablenote{\textsuperscript{a}Analyzed on their own; not included in the decision accuracy.}
\end{table*}

Up to six tasks were constructed for each evaluation patient (\hyperref[tab:1]{Table 1}), namely routine planning with the default goals, a physician note that contradicted a formal goal, a goal on a structure that the case did not contain, and three tasks that tested whether a system could distinguish an impossible request from a difficult one.

In the coverage/cap contradiction, the target D99 goal was raised to 78.0 or 78.5~Gy, above the 77~Gy limit on the maximum dose to the External (body) contour, a conflict evident from the two stated values. In the overlap contradiction, a high target dose was required in tissue shared with the mandible, whose goal allowed too little high dose, so that each shared voxel either underdosed the target or overdosed the mandible and the conflict had to be worked out from the anatomy. Thus, if $O$ voxels are shared and the goals allow at most $m$ underdosed target voxels and $n$ overdosed mandible voxels, $O > m + n$ proves that no plan can meet both goals.

The third task, a feasible control, was difficult but achievable, in that its spinal-cord limit was tightened to a value that a stored plan, found with the same optimizer and budget, had met together with every other hard goal. Tasks that could not be constructed or verified were excluded (\hyperref[tab:1]{Table 1}), and every proof and stored plan was confirmed by independent reconstruction (Supplementary \hyperref[sec:S5]{Section S5}).

\subsection{Planning systems}\label{sec:2.3}\nobreak

The four systems differed in how much of the planning was decided by an LLM (\hyperref[fig:1]{Fig. 1(b)}). The first, the controller, was a rule-based planning script that set objectives from the goals, adjusted their priorities after each optimization, and normalized the plan. The second added hand-written arithmetic checks for missing structures and the two contradictions before passing a task to the controller, and thus indicates what is achievable when a conflict is known in advance. In the third, the interpreter, an LLM reviewed the request once and either escalated it or passed it to the controller, and in the fourth, the full agent, the LLM performed every planning step itself.

Both LLM-based systems were given the same planning policy, namely to follow the formal goals when a note contradicted them, to escalate a goal on a missing structure, to escalate a request as infeasible once an incompatibility between hard goals had been demonstrated, and otherwise to plan within the budget and report any unmet goals. Each was run twice per task with each of two LLMs, Terra (openai/gpt-5.6-terra), a large commercial model accessed online, and Ministral 3 8B (Mistral AI, Paris, France), a smaller, openly available model run on a desktop workstation (Supplementary Sections \hyperref[sec:S1.4]{S1.4} and \hyperref[sec:S9.3]{S9.3}). The scripted systems were run once per task, and the controller was rerun on the feasible controls with 2,000 and 5,000 iterations per optimization.

\subsection{Outcomes}\label{sec:2.4}\nobreak

Four outcomes were recorded for every attempt (\hyperref[fig:1]{Fig. 1(c)}). The first was the final decision, that is, whether the system submitted a plan or escalated the task, and decision accuracy was the percentage of attempts with the expected decision on the four scored tasks (\hyperref[tab:1]{Table 1}). A submission on a contradiction received no credit even when it listed the unmet goals, given that no plan can meet them, whereas a submission on a feasible control received credit regardless of its dose. Decision accuracy was averaged within each patient with patients weighted equally, and 95\% confidence intervals (CIs) were estimated from 10,000 bootstrap resamples of the patients.

The second outcome, dose-goal attainment, was assessed from the dose of each submitted plan, recomputed from its fluence, against the hard goals on target D99, organ D0.1cc, and the External maximum dose (Supplementary \hyperref[tab:S15]{Table S15}). The third, recognition, was whether the explanation that accompanied a decision identified the actual problem; it was read for every overlap attempt and rated by a medical physicist in a sample of 60 responses (Supplementary \hyperref[sec:S3]{Section S3}). The fourth comprised technical failures, such as software errors, replies that did not follow the required format, and attempts ended by the harness, which received no credit and were not repeated.

\subsection{Planning logs of a clinical agent}\label{sec:2.5}\nobreak

The same outcomes were obtained from the planning logs of SAGE~\cite{nusrat2025}, which planned the SRS cases examined here in Eclipse (Varian Medical Systems, Palo Alto, CA) using the QwQ-32B reasoning LLM in a series of numbered attempts. In each attempt, the LLM is given the current value of every goal and the results of up to three previous attempts, states which goals fail, and proposes new objective priorities, with which Eclipse optimizes the plan and calculates the dose; the run ends once every goal passes, or otherwise after 10 attempts.

All 39 SAGE runs in the available records were examined, from 31 patients prescribed 21~Gy in one fraction and planned between July and September 2026, including two refinement runs restarted after an earlier run had not met every goal. The final decision was taken from how each run ended, dose-goal attainment from the goal values after each dose calculation, recognition from whether the failing goals named by the LLM matched those it had been shown, and technical failures from whether each proposal reached Eclipse and produced a dose (Supplementary \hyperref[sec:S7]{Section S7}). This retrospective analysis was approved by the institutional review board (protocol 18702), which waived the requirement for informed consent.

\section{Results}\label{sec:3}\nobreak

\subsection{Decision accuracy}\label{sec:3.1}\nobreak

\begin{figure*}[tp]
\centering
\includegraphics[width=178mm]{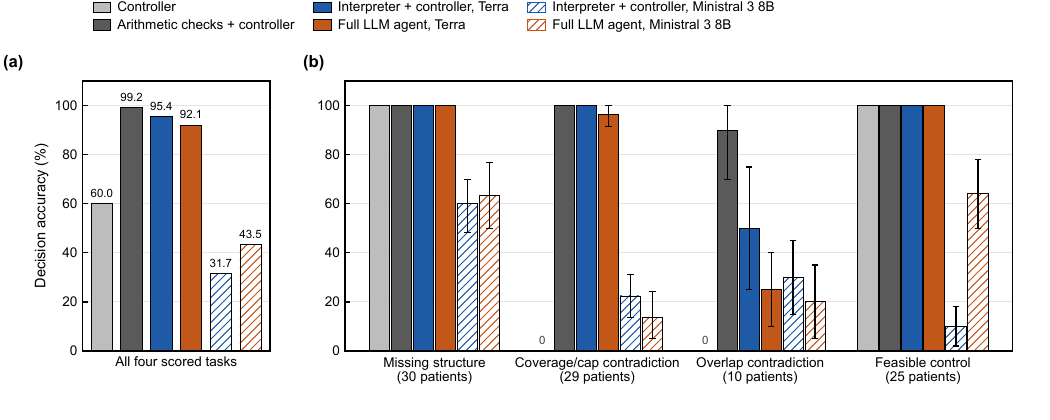}
\caption{Decision accuracy on the 30 evaluation patients. (a) Decision accuracy over the four scored tasks, averaged within each patient with patients weighted equally. (b) Decision accuracy on each scored task, with the number of eligible patients; error bars show 95\% patient-bootstrap confidence intervals, and 0 marks a system that never made the expected decision. The LLM-based systems were run with Terra (solid bars) and with Ministral 3 8B (hatched bars), and technical failures, including replies that did not follow the required format, received no credit.}\label{fig:2}
\end{figure*}

As shown in \hyperref[fig:2]{Fig. 2}, with Terra, the decision accuracy was 95.4\% for the interpreter and 92.1\% for the full agent, compared with 99.2\% for the arithmetic checks and 60.0\% for the controller alone. Both LLM-based systems recognized the problems stated in the request, escalating every missing-structure task and nearly every coverage/cap contradiction (100\% and 96.6\%, respectively) and flagging the contradicting note or escalating in 60 and 59 of 60 attempts. On the overlap contradictions, however, which had to be worked out from the anatomy, the interpreter escalated only 50\% (95\% CI, 25--75\%) of attempts and the full agent only 25\% (10--40\%), whereas the arithmetic checks, working from the same case summary, escalated 9 of the 10.

\begin{figure*}[tp]
\centering
\includegraphics[width=178mm]{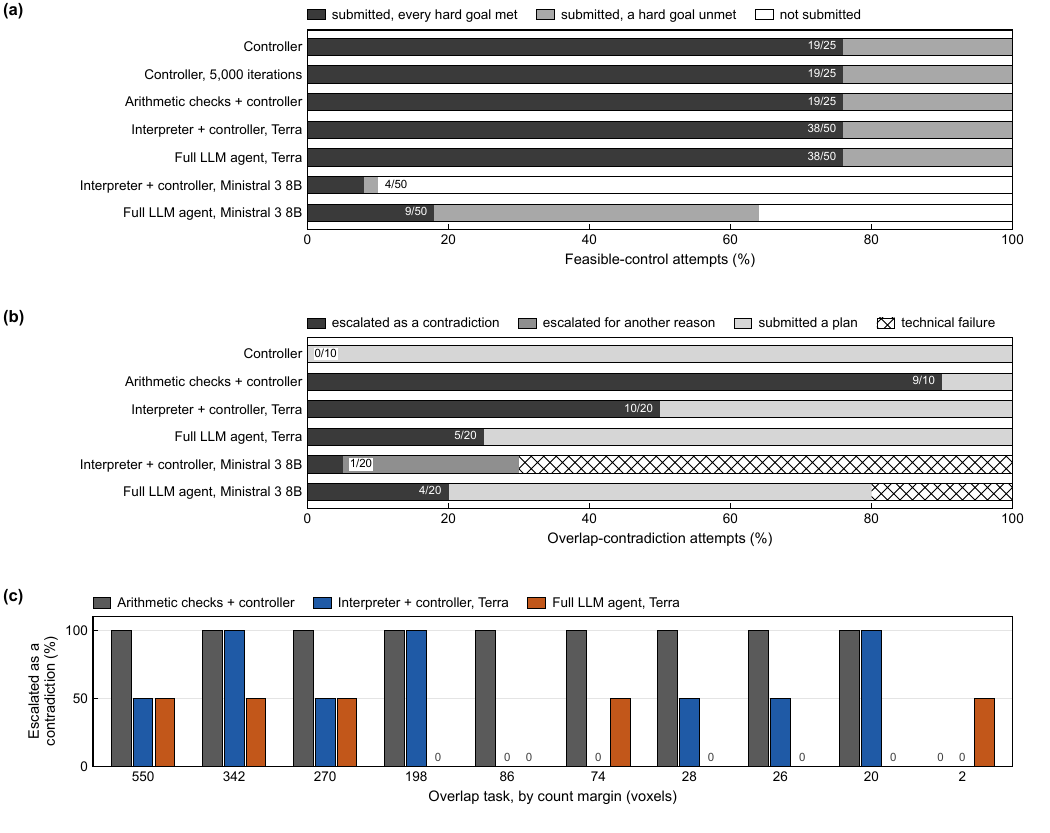}
\caption{Feasible controls and overlap contradictions. (a) Outcome of every feasible-control attempt: submitted with every hard dose goal met, submitted with a hard goal unmet, or not submitted, with the number that met every hard goal. The LLM-based systems made two attempts per control, systems that passed a task to the controller submitted its plans, and the controller with 2,000 iterations gave the same result as with 5,000 (Supplementary \hyperref[tab:S17]{Table S17}). (b) Final decision in every attempt on the 10 overlap contradictions, with the number escalated as a contradiction. (c) Percentage of attempts escalated as a contradiction on each overlap task, with the tasks ordered by how far the shared voxels exceeded the combined allowance (count margin); 0 marks no such escalation.}\label{fig:3}
\end{figure*}

Even on the nine tasks whose shared voxels exceeded the combined allowance by at least 20 voxels, the interpreter missed at least one of its two runs on six and the full agent on all nine (\hyperref[fig:3]{Fig. 3(c)}). When the interpreter passed an overlap task to the controller, its explanation stated that no incompatibility had been demonstrated, and in each of its 15 misses, the full agent used all three optimizations and then submitted a plan listing the unmet goals without describing the request as infeasible. The physicist's review agreed with the automatic credit in all 40 sampled responses from the scored tasks (Supplementary \hyperref[sec:S3]{Section S3}).

Ministral 3 8B missed the same contradictions, with its interpreter and full agent escalating 6 and 4 of the 20 overlap attempts and reaching decision accuracies of only 31.7\% and 43.5\%, respectively; the interpreter's replies did not follow the required format in 113 of 188 attempts, the full agent lost 26 of 188 attempts to planning actions that the local software could not read, and the two systems escalated 8 and 7 of the 50 feasible controls (Supplementary \hyperref[sec:S9.3]{Section S9.3}).

\subsection{Dose-goal attainment on the feasible controls}\label{sec:3.2}\nobreak

As shown in \hyperref[fig:3]{Fig. 3(a)}, the scripted systems and both LLM-based systems with Terra submitted every feasible control; however, only 76\% of these plans met every hard goal, comprising 19 of 25 attempts for each scripted system and 38 of 50 for each LLM-based system. The unmet goals were mainly the D99 of the 70~Gy planning target volume, short by 0.2--9.7~Gy beyond the goal tolerance, and the External maximum dose, exceeded by 1.3--3.5~Gy, whereas the stored plans met the same goals with margins of 2.9--3.4~Gy and 0.4--2.1~Gy, respectively. Every deficient plan was submitted with its unmet goals listed and thus counted as a correct decision; had credit also required every hard goal to be met, each system would have scored 6.4 percentage points lower (Supplementary \hyperref[tab:S14]{Table S14}).

A larger optimization budget allowed the optimizations to converge but did not correct these plans, in that with 2,000 or 5,000 iterations, the same six controls remained deficient with the same unmet goals, and their target D99 changed by less than 0.02~Gy (Supplementary \hyperref[sec:S9.2]{Section S9.2} and \hyperref[tab:S17]{Table S17}). Because the stored plans had met every goal within the original budget, the shortfall was caused by the objectives that the controller selected.

\subsection{The SAGE planning logs}\label{sec:3.3}\nobreak

\begin{figure*}[tp]
\centering
\includegraphics[width=178mm]{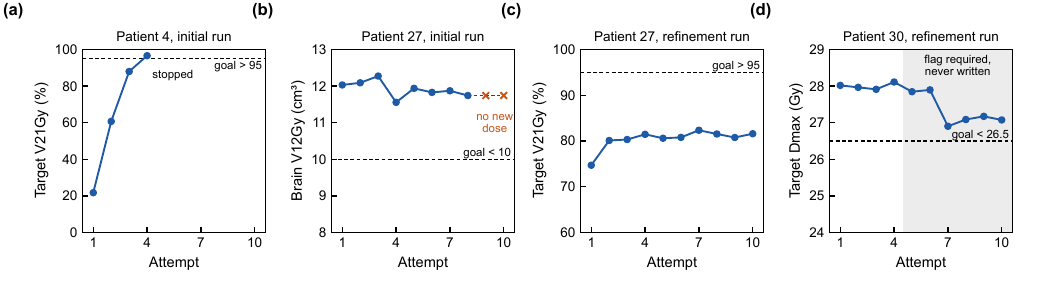}
\caption{The SAGE planning logs, read with the outcomes of OpenGray: measured value of the failing goal at each attempt in (a) a run that met every goal and (b--d) three of the four runs that reached the 10-attempt limit. Dashed lines mark the strict goal thresholds. In (b), crosses mark the goal values of attempt 8, which were reported again after the two attempts in which no dose was calculated. In (d), the shaded region marks the attempts at which the instructions of the run required the flag ``not feasible with priorities,'' which was never written.}\label{fig:4}
\end{figure*}

Of the 39 SAGE runs, 35 met every goal, 28 of them after a single dose calculation. In each of the other four, one goal failed at every attempt until all 10 attempts had been used (\hyperref[fig:4]{Fig. 4}), namely normal-brain V12Gy in the initial run of Patient 27 (11.56--12.27~cm\textsuperscript{3}, against a limit of 10~cm\textsuperscript{3}), target coverage in its refinement run (74.70--82.35\%, against a goal of 95\%), and the target maximum dose in both runs of Patient 30 (26.90--28.24~Gy, against a limit of 26.5~Gy).

The agent recognized its failing goals consistently: in all 88 of its replies, the goals it identified as failing were exactly those that failed in the values it had been shown. However, its instructions provided no action for stopping or requesting review, and no log recorded a stop, an escalation, or a manual intervention. The refinement instructions for Patient 30 did ask for the flag ``not feasible with priorities'' if the target maximum dose still failed after four attempts. The dose failed at all 10 attempts, yet the flag was never written. The LLM was shown only the three most recent attempts and never the attempt number, and in attempt 7, its reasoning counted the three attempts that it could see and deferred the flag.

The logs also revealed a technical failure that a review of the final goal values would have missed. In attempt 9 of the initial run of Patient 27, the reply abbreviated its list of seven objectives, and only the first, an upper limit on the optic chiasm, reached Eclipse, so no dose was calculated in attempts 9 and 10 (\hyperref[fig:4]{Fig. 4(b)} and Supplementary \hyperref[sec:S7]{Section S7}).

\section{Discussion}\label{sec:4}\nobreak

This work focused on evaluating the decisions of radiotherapy planning agents on tasks with known answers and in clinical planning logs. With Terra, both LLM-based systems exceeded 90\% decision accuracy, yet they missed most of the overlap contradictions, and every system submitted feasible plans that missed hard goals. Neither failure was limited to one LLM or explained by the optimization budget, and both became apparent only when the decisions were examined by task and alongside the achieved dose.

This pattern can be explained by the reasoning that each contradiction required. The LLM-based systems recognized contradictions written in the request but missed the one that required arithmetic on the anatomy, although the case summary contained the necessary numbers. Their instructions also allowed an infeasibility escalation only once an incompatibility had been demonstrated and otherwise asked for a plan that reported its unmet goals, the fallback that the full agent took in every miss. Checks written for a known conflict were effective, as rule-based knowledge has been in learned planning~\cite{shen2021}; however, such checks cover only the conflicts that were anticipated, and once the tasks of a benchmark are public, a system can be tuned to them, so future versions of OpenGray will require conflict types withheld from developers.

The aggregate score must also be interpreted with care. Given that the tasks were averaged within each patient, the missing-structure and feasible-control tasks made up 60\% of the score and the overlap task only 9\%, and the controller scored 60.0\% while missing every contradiction. Similarly, credit for submitting a feasible plan says nothing about its dose, so decision accuracy should be reported by task and together with the achieved dose.

The SAGE logs showed a related gap in clinical planning, in that the agent recognized its failing goal in every reply but had no action other than another attempt, and a flag intended for such runs depended on an attempt count that the agent was never given. The full agent behaved similarly on the benchmark, reporting its unmet goals instead of escalating, although an escalation was available to it. Recognizing a failing goal is thus insufficient; an agent also requires an explicit action for stopping and requesting review, a rule for when to take that action, and the information on which the rule depends.

There are several limitations to this work. The benchmark used one head-and-neck dataset, two contradiction types, and 30 evaluation patients, so the results describe these tasks and the two LLMs tested. The feasible controls tightened only the spinal-cord limit and did not include achievable versions of the two contradictions just inside the boundary, so whether a system can tell such a request from a contradiction was not tested. The plans were optimized with no machine constraints, and none of the 540 submitted plans met the fluence-complexity limit of the OpenKBP-Opt reference plans; deliverability constraints would restrict the achievable dose, so the feasible controls would need to be verified again under them. Finally, the log audit covered one agent at one institution and relied on the dose calculated by the planning system, and because no verified answer exists for a clinical case, it cannot show whether the four unresolved runs could have met their goals.

In conclusion, favorable aggregate scores concealed two decision failures that persisted with a second LLM and a larger optimization budget, and a clinical agent recognized a failing goal but had no means of stopping. Any agent that can call the nine planning functions can be tested on OpenGray, and any agent that logs its proposals, their transfer to the planning system, and the resulting dose can be audited with the same four outcomes. We thus recommend that planning agents be 1) evaluated on their decisions as well as their dose, on tasks with known answers before clinical use and through their planning logs afterward, and 2) given an explicit action for stopping and requesting review.

\begingroup\small
\closingsection{Declaration of generative AI and AI-assisted technologies in the manuscript preparation process}
During the preparation of this work the authors used ChatGPT Codex (OpenAI) in order to assist with writing and editing this manuscript and with checking its references. After using this tool, the authors reviewed and edited the content as needed and take full responsibility for the content of the published article.

\closingsection{CRediT authorship contribution statement}
Humza Nusrat: Conceptualization, Methodology, Formal analysis, Writing -- original draft, Writing -- review \& editing. Kundan Thind: Conceptualization, Methodology, Writing -- original draft, Writing -- review \& editing.

\closingsection{Funding}
The authors have no funding sources to disclose for this work.

\closingsection{Declaration of competing interest}
The authors declare that they have no known competing financial interests or personal relationships that could have appeared to influence the work reported in this paper.

\closingsection{Ethics approval}
This retrospective secondary analysis of the SAGE planning records was approved by the Henry Ford Health Institutional Review Board (protocol 18702), which waived the requirement for informed consent.

\closingsection{Data availability statement}
The OpenKBP and OpenKBP-Opt datasets used in this study are publicly available~\cite{babier2021,babier2022}. The OpenGray software and the benchmark data supporting this study can be found in Zenodo (\href{https://doi.org/10.5281/zenodo.23681336}{\nolinkurl{https://doi.org/10.5281/zenodo.23681336}}). The clinical logs and patient-to-plan linkage records are retained privately and are excluded from the deposit.
\par\endgroup

\renewcommand{\bibsection}{\section*{\refname}}
\renewcommand{\refname}{References}

\clearpage
\onecolumngrid
\begin{center}
{\large\bfseries Supplementary material\par}
\vspace{6pt}
{\itshape OpenGray: a benchmark for evaluating the decisions of radiotherapy planning agents\par}
\end{center}
\vspace{10pt}
\twocolumngrid
\pdfbookmark[0]{Supplementary material}{supplement}
\setcounter{section}{0}\renewcommand{\thesection}{S\arabic{section}}\renewcommand{\theHsection}{supp.\arabic{section}}
\setcounter{table}{0}\renewcommand{\thetable}{S\arabic{table}}\renewcommand{\theHtable}{supp.\arabic{table}}
\setcounter{figure}{0}\renewcommand{\thefigure}{S\arabic{figure}}\renewcommand{\theHfigure}{supp.\arabic{figure}}
\renewcommand{\figurename}{Figure}
\makeatletter\def\@seccntformat#1{\csname the#1\endcsname\quad}\makeatother

\section{Planning environment, task definitions, and execution}\label{sec:S1}\nobreak

\subsection{Planning interface and task policy}\label{sec:S1.1}\nobreak

OpenGray exposes nine planning functions, which read the case summary, set objectives, optimize the fluence, obtain dose-volume histograms, obtain dose metrics, compare the dose with the goals, normalize the dose, submit a plan, and escalate a task. Every task is run by the benchmark harness, the software that presents the case to a planning system, carries out each function that the system requests, and records every step and the resulting dose. The development experiments were organized in five tracks, T1--T5, and the reserved evaluation used the six tasks of \hyperref[tab:1]{Table 1} of the main manuscript.

T1 permits one optimization, T2 permits iterative planning, and T3 introduces a tightened or relaxed goal immediately after the first optimization. T4 includes four changes to the presentation of the case, conflicting physician instructions, a missing required structure, and an unaltered control; a further scenario with a tightened spinal-cord limit had no proof of infeasibility and was excluded from the T4 comparisons. T5 includes coverage/cap contradictions, target--organ overlap contradictions, and difficult feasible controls. The certificate definitions and construction exclusions are given in \hyperref[sec:S5]{Section S5}.

In development, a T4 response to a conflicting note is labeled correct when it escalates or submits a plan that explicitly identifies the note. The T4 grade of a submission is its PlanScore whether or not the note is identified, so the grade does not penalize a system that ignores the note. For certified T5 contradictions, the accepted escalation reasons are infeasible and contradictory instructions. In both tracks, an appropriate escalation receives a grade of 0.9, and a submission on a certified contradiction receives its PlanScore minus 0.25, with a floor of zero. Thus, an escalation count, a track grade, and recognition of the problem measure different outcomes. The reserved evaluation used the decision accuracy defined in the main manuscript.

\subsection{Dose goals and PlanScore}\label{sec:S1.2}\nobreak

PlanScore combines hard-goal attainment, soft-goal violation, and sparing relative to the reference mean doses as
\[
S=0.5H+0.3(1-V)+0.2R.
\]
Here, $H$ is the fraction of hard goals met and $V$ is the mean fractional violation of the soft goals. Upper dose goals allow the nominal limit plus 0.05~Gy, and lower goals allow the nominal limit minus 0.05~Gy; this 0.05~Gy tolerance applies to both hard and soft dose-valued goals. A fractional violation is the excess beyond the tolerated boundary divided by the nominal limit and clipped to [0,1], and a goal that is met contributes zero violation.

The reference component includes the soft mean-dose goals of non-target structures with positive reference mean doses. For each eligible structure, the fractional reduction is the reference mean dose minus the achieved mean dose, divided by the reference mean dose; these values are clipped to [\textminus{}1,1], averaged, and mapped to [0,1]. Thus, $R$ rewards sparing relative to the reference, and it equals 0.5 when no eligible reference values exist. In the absence of hard goals, $H=1$, and in the absence of soft goals, $V=0$.

Two rejection gates set PlanScore to zero if the global maximum dose exceeds 115\% of the highest prescription or if any target D99 falls below 80\% of its prescription. A visible External maximum-dose goal is also set at 110\% of the highest prescription. Missing any other hard goal reduces $H$ but does not reject the plan automatically. As such, PlanScore represents the specified weighting of the criteria and imposes no strict clinical priority order. Across the 2,159 successful T1, T2, and evaluable T4 development episodes, the ranking of the systems changed little with a greater hard-goal weight (Kendall rank correlations with the baseline of 0.857--0.970) but more with the rejection gate tightened to 110\% (0.515--0.786).

\subsection{Development systems and comparison schedules}\label{sec:S1.3}\nobreak

The 10 development patients were pt\_258, pt\_269, pt\_289, pt\_302, pt\_306, pt\_307, pt\_320, pt\_324, pt\_329, and pt\_339 within the OpenKBP-Opt dataset. The scripted comparators included a priority-escalation heuristic, a controller with objective-level back-off and rule-based normalization, random search, and Optuna search with access to PlanScore. The preflight variant added a missing-structure check to the controller, and the arithmetic variant also checked the two constructed contradiction families. The design of the controller incorporated observations from the trajectories of development models, and the arithmetic checks were designed for the certified constructions.

The full agents used the stored routes openai/gpt-5.6-terra, anthropic/claude-sonnet-5, and z-ai/glm-5.3-flash, abbreviated Terra, Sonnet, and GLM, respectively; two further models, openai/gpt-oss-120b and google/gemini-3.1-flash-lite, were run only in early T1 and T2 runs with the rules withheld and were not carried into the comparisons. The interpreter used one Terra call to either escalate the task or delegate planning to the controller. Its reduced summary contained the structure names and volumes, prescriptions, goals, notes, and rejection rules, while the full summary added anatomical fields (including overlap information and voxel counts) and beam geometry. The interpreter prompt and model route were identical under both summary conditions.

The rejection-rule experiment crossed three models with disclosed and withheld conditions on all 10 patients, with five episodes per condition. The Terra T2 results come from a rerun with the rules disclosed, whereas the Sonnet and GLM T2 results retain the historical conditions with the rules withheld; as such, they cannot be used to rank these models under common disclosure conditions. T3 used five repetitions per system and patient, with tightened and relaxed updates. The scripted T4 controls covered the complete patient-by-scenario set, while the language agents received rotated scenarios with six or seven episodes per scenario. T5 used five repetitions for the language systems and one per patient-scenario for the deterministic controls.

Repetitions were averaged within each patient, and patients were weighted equally. Comparisons of rotated scenarios first matched the patient and scenario and then averaged the task differences within each patient. The development confidence intervals used 2,000 patient-bootstrap resamples without adjustment for multiplicity, and these comparisons are exploratory. When an episode had several attempt records, the newest successful record was kept, and an error was recorded only when no attempt succeeded; such errors were reported apart from the dose outcomes.

The complexity experiment added fluence penalties and a maximum beamlet-weight cap to approximate the conditions of the reference optimization, and the subsequent normalization could violate these limits.

\subsection{Reserved-cohort protocol and execution}\label{sec:S1.4}\nobreak

The reserved evaluation compared the controller, arithmetic checks plus controller, interpreter plus controller, and full agent on 30 patients. Routine, missing-structure, and conflicting-note tasks were available for all patients, and coverage/cap contradictions could be constructed for 29 patients, overlap contradictions for 10, and feasible controls for 25. This yielded 154 eligible tasks per system. The complete schedule contained 1,080 entries, comprising 924 eligible attempts and 156 construction exclusions, and the four primary families contained 94 eligible patient-family tasks and 564 attempts across systems and repetitions.

Both language systems used the stored Terra route and the full task-relevant summary. The interpreter was limited to one model call and 2,048 output tokens, and the full agent could use up to 15 model calls and 8,192 output tokens per call. Both systems were allowed three optimizations and 15 planning actions per episode. The deterministic systems ran once per eligible task, and the language systems had two fresh repetitions with provider seeds 0 and 1. Temperature was omitted because the endpoint did not support it, and response caching, automatic retries, and provider fallback were disabled. The route identifiers specify the requested endpoints, and the versions of the serving models could not be verified independently. \hyperref[tab:S15]{Table S15} specifies the planning environment, and \hyperref[tab:S16]{Table S16} gives the execution settings.

Before any candidate was evaluated, the task inputs, source code, prompts, patient-cache identities, feasible witnesses, analysis, and response-review selection were checksummed and committed privately (\hyperref[tab:S16]{Table S16}), and hashes fixed the order of patients and episodes. The primary paired patient bootstrap used 10,000 resamples and seed 20260909.

The primary policy required escalation for missing required structures and certified contradictions, and explicit submission for feasible controls; the percentage of attempts in which a system took the required action is reported as its decision accuracy. Routine planning and conflicting instructions were secondary scenarios. In the conflicting-note tasks, the structured goals took precedence, and an acceptable response identified the conflict and then either escalated or submitted according to the formal goals. Errors, automatic submissions, and endings forced by the harness received zero primary credit. Repetitions were averaged within each patient and family, the eligible families were then averaged within each patient, and patients were weighted equally. The prespecified contrast was the interpreter/controller minus the full agent, with a 95\% percentile interval from the paired bootstrap. The exact paired sign-flip test of this contrast, the runs of both language systems with Ministral 3 8B, the controller runs with longer optimization budgets, and the further analyses in Sections \hyperref[sec:S8]{S8} and \hyperref[sec:S9]{S9} were added after the reserved evaluation, and all other comparisons between systems are descriptive.

All scheduled attempts and construction exclusions were retained. The campaign was interrupted twice and was resumed only for attempts that had not yet started, so no failed or unfinished attempt was replaced. Six attempts failed for technical reasons: two connection failures, one model-request timeout, and three interrupted attempts of the deterministic systems. \hyperref[tab:S8]{Table S8} lists each failure and whether it belonged to the primary endpoint.

\section{Development results}\label{sec:S2}\nobreak

\subsection{Rejection-rule disclosure and planning performance}\label{sec:S2.1}\nobreak

Disclosing the rejection rules increased the T1 PlanScore by 0.281 for Terra (95\% confidence interval [CI], 0.099--0.472), 0.194 for Sonnet (0.020--0.402), and 0.146 for GLM (0.016--0.306). Among successful episodes, rejection-gate failures decreased from 46/150 with the rules withheld to 7/149 with the rules disclosed. One attempt under disclosure ended with a provider error and was excluded from the dose outcomes.

With the rules disclosed, the mean T1 PlanScore was 0.864 for the controller, 0.847 for Terra, 0.811 for GLM, and 0.785 for Sonnet (\hyperref[fig:S4]{Figure S4}). On T2, Terra and the controller scored 0.878 and 0.879, respectively, with a paired difference of \textminus{}0.001 (95\% CI, \textminus{}0.027 to 0.026) that does not demonstrate equivalence. Under the historical conditions with the rules withheld, the T2 scores of Sonnet and GLM were 0.791 and 0.606, respectively.

\subsection{Response to revised goals}\label{sec:S2.2}\nobreak

All 300 selected T3 episodes recorded exposure to the revised goal after the first optimization. The mean scores were 0.887 for Terra, 0.876 for the controller and interpreter, 0.872 for Sonnet, 0.861 for GLM, and 0.855 for the heuristic. The matched differences relative to the heuristic were 0.032 for Terra (95\% CI, 0.011--0.054) and 0.022 for the controller (0.003--0.044). The tightened goal was attained in 30/30 controller and interpreter episodes, 29/30 Terra episodes, 27/30 Sonnet and GLM episodes, and 24/30 heuristic episodes.

\subsection{Altered inputs and explicit checks}\label{sec:S2.3}\nobreak

\begin{figure*}[tp]
\centering
\includegraphics[width=178mm]{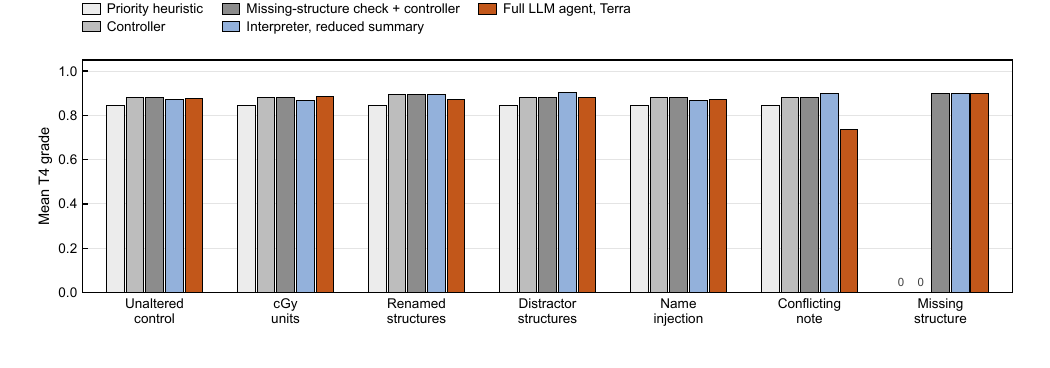}
\caption{Development T4 grade by evaluable scenario: mean grade of each system in each scenario. Scripted controls cover 10 patients per scenario (the priority heuristic, 50 episodes), and the rotated language-system scenarios contain six or seven episodes. An appropriate escalation earns 0.9, so the grade combines action credit and dose outcomes, and 0 marks a grade of zero. The scripted systems ignore the note, and a submission is graded by its PlanScore whether or not the note is flagged, so their conflicting-note grades equal their grades on the unaltered control. The tightened-cord scenario, which had no proof of infeasibility, is excluded.}\label{fig:S1}
\end{figure*}

On the evaluable T4 scenarios (\hyperref[fig:S1]{Figure S1}), the matched grade differences relative to the heuristic were 0.161 for the interpreter (95\% CI, 0.103--0.215), 0.159 for preflight (0.131--0.205), 0.139 for the full Terra agent (0.088--0.187), and 0.030 for the controller (0.002--0.077). Adding the missing-structure rule increased the grade of the controller by 0.129 across its seven-scenario population, and this increase arose from escalation credit and did not reflect improved dose quality.

The interpreter escalated on all six conflicting-note episodes and all six missing-structure episodes. The full Terra agent also escalated on all six missing-structure episodes; however, it submitted on all six conflicting-note episodes, and the automatic grader identified no conflict flag in these submissions. The development review credited recognition of the problem in 6 of 12 sampled conflicting-note responses and in all 12 missing-structure responses. The arithmetic variant reproduced the behavior of the preflight variant on the evaluable T4 scenarios.

\subsection{Anatomical information and contradiction recognition}\label{sec:S2.4}\nobreak

With the reduced summary, the interpreter escalated on 50/50 coverage/cap contradictions and 0/25 overlap contradictions, and with the full summary, it escalated on 49/50 and 25/25, respectively. The reduced-summary overlap episodes, however, came from an earlier version of the case summary, in which the goal note was not updated when the organ limit changed, and the full summary also added beam geometry, so this difference cannot be attributed to the anatomical fields alone. Neither condition led to escalation on any of the 50 feasible controls. The full LLM agents escalated on 27--49 of 50 coverage/cap episodes and 0--5 of 25 overlap episodes. These counts include escalations for any reason, and thus the correctness of the explanations cannot be judged from them.

\begin{figure*}[tp]
\centering
\includegraphics[width=178mm]{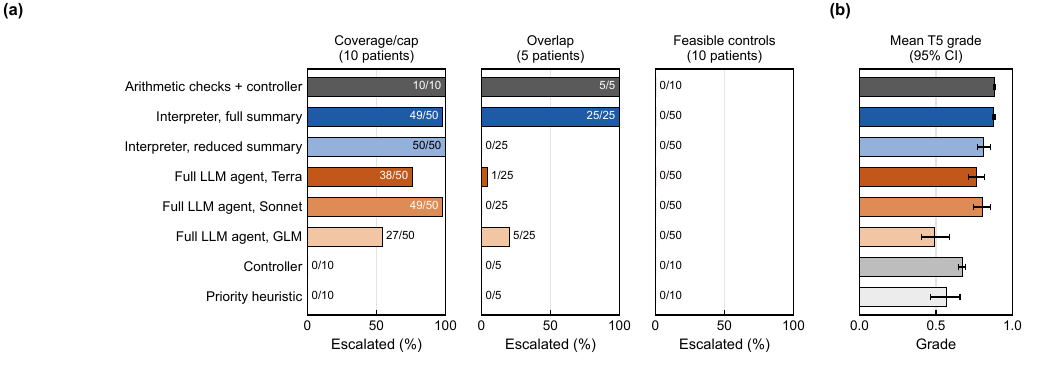}
\caption{Development T5 escalation frequency and grade. (a) Percentage of episodes escalated for any reason in each task, with the counts beside the bars; coverage/cap contradictions involve 10 patients and overlap contradictions 5, and no system escalated on a feasible control. (b) Patient-mean track grade with 95\% patient-bootstrap confidence intervals.}\label{fig:S2}
\end{figure*}

The arithmetic checks escalated all 15 certified development patient-scenarios and none of the 10 feasible controls, and their mean T5 grade was 0.882 (95\% CI, 0.874--0.890), compared with 0.878 (0.870--0.886) for the full-summary interpreter. \hyperref[fig:S2]{Figure S2} shows the escalation counts and track grades of every system.

\subsection{Fluence-complexity objectives}\label{sec:S2.5}\nobreak

\begin{figure*}[tp]
\centering
\includegraphics[width=178mm]{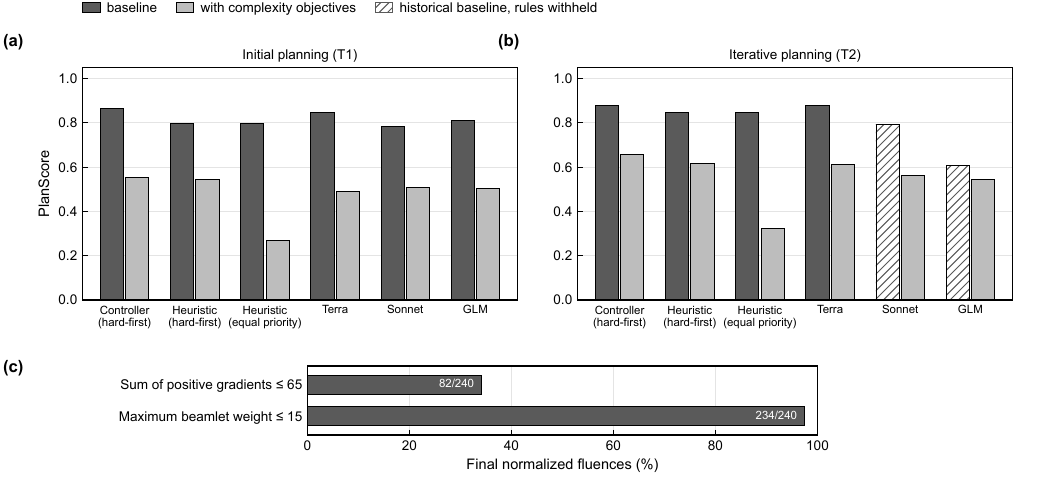}
\caption{Sensitivity to fluence-complexity objectives. (a, b) Patient-mean PlanScore without (dark) and with (light) the added complexity objectives in T1 and T2; hatched bars are the historical Sonnet and GLM T2 baselines with the rules withheld. (c) Percentage of the 240 final normalized fluences that met the intended limits, which normalization could violate.}\label{fig:S3}
\end{figure*}

\begin{figure*}[tp]
\centering
\includegraphics[width=178mm]{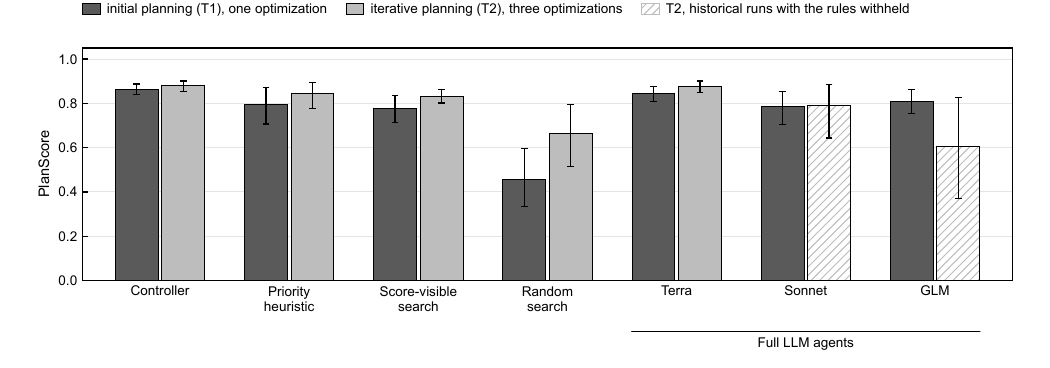}
\caption{Planning performance on the 10 development patients: patient-mean PlanScore in initial planning (T1; one optimization, rules disclosed) and iterative planning (T2; three optimizations), with 95\% patient-bootstrap confidence intervals. Hatched bars are the historical Sonnet and GLM T2 runs with the rules withheld, which are not directly comparable with the others. The score-visible search had access to PlanScore during planning.}\label{fig:S4}
\end{figure*}

In 240 complexity experiments (\hyperref[fig:S3]{Figure S3}), the mean PlanScore decreased by 0.061--0.528 relative to the corresponding development results without the added complexity objectives, and a fixed priority order improved the constrained heuristic relative to equal-priority objectives. However, only 82/240 final normalized fluences met the intended sum-of-positive-gradients limit of 65, and 234/240 met the maximum beamlet-weight limit of 15, since normalization could violate both limits. The changes in score thus reflect the added objectives and the planning choices they induced, and they do not demonstrate clinical deliverability.

\section{Review of agent responses}\label{sec:S3}\nobreak

\subsection{Sampling and review procedure}\label{sec:S3.1}\nobreak

The review included 138 development responses, comprising 58 coverage/cap, 46 overlap, 12 conflicting-note, 12 missing-structure, and 10 feasible-control responses. The reserved sample included 10 responses per scenario, 15 per system, and distinct patients within each scenario; recognition was not applicable to routine planning or feasible controls. The stratified samples were designed to identify errors, and they were not intended to estimate the prevalence of those errors.

One reviewer, a medical physicist, checked every source response and assigned its action and recognition ratings through an offline interface, with system identities and automatic grades withheld.

The development rubric credited an escalation recorded on a contradiction task, including nine escalations forced by the harness, whereas the reserved rubric gave such endings no credit.

\subsection{Ratings}\label{sec:S3.2}\nobreak

\begin{table*}[tp]
\caption{Reviewed responses by task and cohort. Counts use the cohort-specific rubrics, and these stratified samples do not provide pooled population estimates.}\label{tab:S1}
\small
\setlength{\tabcolsep}{3.0pt}
\begin{tabular*}{\linewidth}{@{\extracolsep{\fill}}p{68.19pt}p{116.73pt}p{43.10pt}p{119.53pt}p{94.93pt}@{}}
\hline\hline
\noalign{\vskip 2pt}
\multicolumn{1}{@{}c}{\parbox[b]{68.19pt}{\raggedright Set\strut}} & \multicolumn{1}{c}{\parbox[b]{116.73pt}{\centering Task\strut}} & \multicolumn{1}{c}{\parbox[b]{43.10pt}{\centering Responses\strut}} & \multicolumn{1}{c}{\parbox[b]{119.53pt}{\centering Recognition yes / applicable\strut}} & \multicolumn{1}{c@{}}{\parbox[b]{94.93pt}{\centering Action yes / responses\strut}} \tabularnewline
\hline
\noalign{\vskip 1pt}
\raggedright Development & \centering Conflicting instructions & \centering 12 & \centering 6/12 & \centering 6/12 \tabularnewline
\raggedright Development & \centering Missing structure & \centering 12 & \centering 12/12 & \centering 12/12 \tabularnewline
\raggedright Development & \centering Coverage/cap contradiction & \centering 58 & \centering 36/58 & \centering 40/58 \tabularnewline
\raggedright Development & \centering Overlap contradiction & \centering 46 & \centering 16/46 & \centering 14/46 \tabularnewline
\raggedright Development & \centering Feasible control & \centering 10 & \centering Not applicable & \centering 10/10 \tabularnewline
\noalign{\vskip 3pt}
\raggedright Reserved cohort & \centering Routine planning & \centering 10 & \centering Not applicable & \centering 10/10 \tabularnewline
\raggedright Reserved cohort & \centering Conflicting instructions & \centering 10 & \centering 4/10 & \centering 4/10 \tabularnewline
\raggedright Reserved cohort & \centering Missing structure & \centering 10 & \centering 10/10 & \centering 10/10 \tabularnewline
\raggedright Reserved cohort & \centering Coverage/cap contradiction & \centering 10 & \centering 8/10 & \centering 8/10 \tabularnewline
\raggedright Reserved cohort & \centering Overlap contradiction & \centering 10 & \centering 3/10 & \centering 3/10 \tabularnewline
\raggedright Reserved cohort & \centering Feasible control & \centering 10 & \centering Not applicable & \centering 10/10 \tabularnewline
\hline\hline
\end{tabular*}
\end{table*}

All 198 responses were rated. In the development sample, recognition was credited on 70/128 applicable responses and action on 82/138 responses, and the reserved counts were 25/40 and 45/60, respectively (\hyperref[tab:S1]{Table S1}). Given the differences in sampling and rubric, these counts cannot be pooled. The action ratings agreed with the automatic endpoint on all 40 reviewed reserved responses from the four primary families, and the development action ratings agreed with the saved automatic labels on 129/138 responses; all nine disagreements were the escalations forced by the harness, which the development rubric credited.

\section{Robustness of the development comparisons}\label{sec:S4}\nobreak

The development comparisons rest on 10 patients, so they were checked in two further ways. First, t intervals from the same patient differences were wider than the bootstrap intervals, and four comparisons whose bootstrap interval excluded zero had a t interval that included it: the T1 disclosure gains of Sonnet (\textminus{}0.040 to 0.428) and GLM (\textminus{}0.032 to 0.325), Sonnet minus the controller on T1 (\textminus{}0.173 to 0.016), and the controller minus the heuristic on T4 (\textminus{}0.020 to 0.081); the disclosure gain of Terra remained positive (0.052--0.510). Second, omitting any one patient left the disclosure gain positive for every model, with ranges of 0.222--0.314 for Terra, 0.121--0.234 for Sonnet, and 0.087--0.182 for GLM, whereas the T2 difference between Terra and the controller ranged from \textminus{}0.0115 to 0.0082 and changed sign. These comparisons are therefore exploratory.

\section{Certificate proofs and feasible controls}\label{sec:S5}\nobreak

\subsection{Dose-volume conventions}\label{sec:S5.1}\nobreak

For a sorted structure-dose vector with $N$ entries and zero-based indexing, a percentile at fraction $q$ interpolates linearly between the order statistics $\operatorname{floor}(q(N-1))$ and $\operatorname{ceil}(q(N-1))$. For ${D}_{p}$, $q=1-p/100$, and for ${D}_{v\mathrm{cc}}$,
\[
k=\max\{1,\operatorname{round}(v/{v}_{\mathrm{voxel}})\},\quad \quad q=1-k/N,
\]
using the nearest-integer rounding of the source implementation. Target evaluation merges the named target with any higher-prescription targets, and source voxels outside the feasible-dose mask contribute zero dose. The checked cases have a uniform voxel volume, External contains every calculated voxel, and both dose and fluence are nonnegative.

\subsection{Overlap contradictions}\label{sec:S5.2}\nobreak

Let $t=0.05$ Gy. For a lower target goal ${D}_{p}\geq c$, we use the conservative allowance
\[
m=\operatorname{ceil}[(1-p/100)(N-1)-{10}^{-9}]+1
\]
for target voxels below $c-t$, which includes an extra voxel to accommodate interpolation. For an upper organ goal ${D}_{v\mathrm{cc}}\leq L$, the conservative hot-voxel allowance is $n=k+1$, and for ${D}_{\mathrm{max}}$, we use $n=1$. If $O$ organ voxels overlap the merged target, the sufficient conditions for incompatibility are
\[
O>m+n,\quad \quad c-t>L+t.
\]
Meeting the target coverage requires at least $O-m$ overlapping voxels at or above $c-t$, while the organ requirement permits at most $n$ voxels above $L+t$, and both requirements cannot hold at the same time under these conditions. The extra rounding allowances make the test sufficient but not necessary, and the proof is independent of optimization performance.

For pt\_248, 631 mandible voxels overlap PTV56, and with $m=286$ and $n=3$, the count margin is 342 voxels. The target lower goal is 53.2~Gy and the organ upper goal is 47.7~Gy; their tolerated boundaries are 53.15 and 47.75~Gy, respectively, giving a dose margin of 5.40~Gy.

\subsection{Coverage/cap contradictions and construction eligibility}\label{sec:S5.3}\nobreak

An External cap ${D}_{\mathrm{max}}\leq E$ bounds every calculated target voxel by $E+t$, and the zeros outside the mask also satisfy that cap. Target coverage requires at least $N-m$ voxels at or above $c-t$, and a sufficient contradiction and its equivalent threshold are
\[
(N-m)(c-t)>N(E+t),\quad \quad c>\frac{N(E+t)}{N-m}+t.
\]
The construction rounds the required coverage upward to a half-gray value, which must remain below 115\% of the highest prescription. For pt\_242, $N=1930$, $m=21$, $E=77$ Gy, and $c=78$ Gy; thus, the exact threshold is 77.9476~Gy and the positive margin is 0.0524~Gy.

\begin{table*}[tp]
\caption{Coverage/cap certificates. D denotes development and R denotes reserved patients. All doses are in Gy. Margins are calculated using unrounded thresholds.}\label{tab:S2}
\small
\setlength{\tabcolsep}{3.0pt}
\begin{tabular*}{\linewidth}{@{\extracolsep{\fill}}p{31.44pt}p{14.24pt}p{24.82pt}p{15.29pt}p{27.60pt}p{46.81pt}p{42.89pt}p{30.93pt}@{}}
\hline\hline
\noalign{\vskip 2pt}
\multicolumn{1}{@{}c}{\parbox[b]{31.44pt}{\raggedright Case\strut}} & \multicolumn{1}{c}{\parbox[b]{14.24pt}{\centering Set\strut}} & \multicolumn{1}{c}{\parbox[b]{24.82pt}{\centering N\strut}} & \multicolumn{1}{c}{\parbox[b]{15.29pt}{\centering m\strut}} & \multicolumn{1}{c}{\parbox[b]{27.60pt}{\centering Cap E\strut}} & \multicolumn{1}{c}{\parbox[b]{46.81pt}{\centering Coverage c\strut}} & \multicolumn{1}{c}{\parbox[b]{42.89pt}{\centering Threshold\strut}} & \multicolumn{1}{c@{}}{\parbox[b]{30.93pt}{\centering Margin\strut}} \tabularnewline
\hline
\noalign{\vskip 1pt}
\raggedright pt\_258 & \centering D & \centering 4635 & \centering 48 & \centering 77.00 & \centering 78.00 & \centering 77.9063 & \centering 0.0937 \tabularnewline
\raggedright pt\_269 & \centering D & \centering 14079 & \centering 142 & \centering 77.00 & \centering 78.00 & \centering 77.8850 & \centering 0.1150 \tabularnewline
\raggedright pt\_289 & \centering D & \centering 4997 & \centering 51 & \centering 77.00 & \centering 78.00 & \centering 77.8945 & \centering 0.1055 \tabularnewline
\raggedright pt\_302 & \centering D & \centering 6440 & \centering 66 & \centering 77.00 & \centering 78.00 & \centering 77.8978 & \centering 0.1022 \tabularnewline
\raggedright pt\_306 & \centering D & \centering 4430 & \centering 46 & \centering 77.00 & \centering 78.00 & \centering 77.9085 & \centering 0.0915 \tabularnewline
\raggedright pt\_307 & \centering D & \centering 4726 & \centering 49 & \centering 77.00 & \centering 78.00 & \centering 77.9072 & \centering 0.0928 \tabularnewline
\raggedright pt\_320 & \centering D & \centering 7887 & \centering 80 & \centering 77.00 & \centering 78.00 & \centering 77.8895 & \centering 0.1105 \tabularnewline
\raggedright pt\_324 & \centering D & \centering 4392 & \centering 45 & \centering 77.00 & \centering 78.00 & \centering 77.8976 & \centering 0.1024 \tabularnewline
\raggedright pt\_329 & \centering D & \centering 497 & \centering 6 & \centering 77.00 & \centering 78.50 & \centering 78.0415 & \centering 0.4585 \tabularnewline
\raggedright pt\_339 & \centering D & \centering 6278 & \centering 64 & \centering 77.00 & \centering 78.00 & \centering 77.8936 & \centering 0.1064 \tabularnewline
\noalign{\vskip 3pt}
\raggedright pt\_242 & \centering R & \centering 1930 & \centering 21 & \centering 77.00 & \centering 78.00 & \centering 77.9476 & \centering 0.0524 \tabularnewline
\raggedright pt\_245 & \centering R & \centering 17522 & \centering 177 & \centering 77.00 & \centering 78.00 & \centering 77.8863 & \centering 0.1137 \tabularnewline
\raggedright pt\_248 & \centering R & \centering 1011 & \centering 12 & \centering 77.00 & \centering 78.50 & \centering 78.0255 & \centering 0.4745 \tabularnewline
\raggedright pt\_249 & \centering R & \centering 9288 & \centering 94 & \centering 77.00 & \centering 78.00 & \centering 77.8878 & \centering 0.1122 \tabularnewline
\raggedright pt\_251 & \centering R & \centering 8299 & \centering 84 & \centering 77.00 & \centering 78.00 & \centering 77.8879 & \centering 0.1121 \tabularnewline
\raggedright pt\_252 & \centering R & \centering 15066 & \centering 152 & \centering 77.00 & \centering 78.00 & \centering 77.8853 & \centering 0.1147 \tabularnewline
\raggedright pt\_256 & \centering R & \centering 3489 & \centering 36 & \centering 77.00 & \centering 78.00 & \centering 77.9033 & \centering 0.0967 \tabularnewline
\raggedright pt\_263 & \centering R & \centering 3213 & \centering 34 & \centering 77.00 & \centering 78.00 & \centering 77.9241 & \centering 0.0759 \tabularnewline
\raggedright pt\_264 & \centering R & \centering 7602 & \centering 78 & \centering 77.00 & \centering 78.00 & \centering 77.8988 & \centering 0.1012 \tabularnewline
\raggedright pt\_267 & \centering R & \centering 1474 & \centering 16 & \centering 77.00 & \centering 78.00 & \centering 77.9455 & \centering 0.0545 \tabularnewline
\raggedright pt\_275 & \centering R & \centering 2656 & \centering 28 & \centering 77.00 & \centering 78.00 & \centering 77.9209 & \centering 0.0791 \tabularnewline
\raggedright pt\_278 & \centering R & \centering 5061 & \centering 52 & \centering 77.00 & \centering 78.00 & \centering 77.8999 & \centering 0.1001 \tabularnewline
\raggedright pt\_282 & \centering R & \centering 9888 & \centering 100 & \centering 77.00 & \centering 78.00 & \centering 77.8872 & \centering 0.1128 \tabularnewline
\raggedright pt\_283 & \centering R & \centering 5139 & \centering 53 & \centering 77.00 & \centering 78.00 & \centering 77.9029 & \centering 0.0971 \tabularnewline
\raggedright pt\_287 & \centering R & \centering 14735 & \centering 149 & \centering 77.00 & \centering 78.00 & \centering 77.8871 & \centering 0.1129 \tabularnewline
\raggedright pt\_292 & \centering R & \centering 9779 & \centering 99 & \centering 77.00 & \centering 78.00 & \centering 77.8880 & \centering 0.1120 \tabularnewline
\raggedright pt\_299 & \centering R & \centering 3361 & \centering 35 & \centering 77.00 & \centering 78.00 & \centering 77.9108 & \centering 0.0892 \tabularnewline
\raggedright pt\_303 & \centering R & \centering 1756 & \centering 19 & \centering 77.00 & \centering 78.00 & \centering 77.9428 & \centering 0.0572 \tabularnewline
\raggedright pt\_304 & \centering R & \centering 6925 & \centering 71 & \centering 77.00 & \centering 78.00 & \centering 77.8982 & \centering 0.1018 \tabularnewline
\raggedright pt\_305 & \centering R & \centering 6041 & \centering 62 & \centering 77.00 & \centering 78.00 & \centering 77.8990 & \centering 0.1010 \tabularnewline
\raggedright pt\_312 & \centering R & \centering 6635 & \centering 68 & \centering 77.00 & \centering 78.00 & \centering 77.8978 & \centering 0.1022 \tabularnewline
\raggedright pt\_313 & \centering R & \centering 8149 & \centering 83 & \centering 77.00 & \centering 78.00 & \centering 77.8929 & \centering 0.1071 \tabularnewline
\raggedright pt\_317 & \centering R & \centering 6972 & \centering 71 & \centering 77.00 & \centering 78.00 & \centering 77.8927 & \centering 0.1073 \tabularnewline
\raggedright pt\_319 & \centering R & \centering 10505 & \centering 107 & \centering 77.00 & \centering 78.00 & \centering 77.8929 & \centering 0.1071 \tabularnewline
\raggedright pt\_323 & \centering R & \centering 35702 & \centering 359 & \centering 77.00 & \centering 78.00 & \centering 77.8826 & \centering 0.1174 \tabularnewline
\raggedright pt\_326 & \centering R & \centering 4337 & \centering 45 & \centering 77.00 & \centering 78.00 & \centering 77.9078 & \centering 0.0922 \tabularnewline
\raggedright pt\_328 & \centering R & \centering 1150 & \centering 13 & \centering 77.00 & \centering 78.50 & \centering 77.9810 & \centering 0.5190 \tabularnewline
\raggedright pt\_334 & \centering R & \centering 5503 & \centering 57 & \centering 77.00 & \centering 78.00 & \centering 77.9064 & \centering 0.0936 \tabularnewline
\raggedright pt\_337 & \centering R & \centering 22295 & \centering 224 & \centering 77.00 & \centering 78.00 & \centering 77.8820 & \centering 0.1180 \tabularnewline
\hline\hline
\end{tabular*}
\end{table*}

Independent reconstruction from the geometry verified all 39 coverage/cap and 15 overlap certificates across the development and reserved cohorts (Tables \hyperref[tab:S2]{S2} and \hyperref[tab:S3]{S3}). The reserved coverage margins were 0.0524--0.5190~Gy, and the overlap count margins were 2--550 voxels. The excluded coverage case, pt\_330, required a threshold of 81.92~Gy, which is above its rejection boundary of 80.50~Gy, and 20 reserved overlap constructions lacked an admissible tightened limit with a positive certificate. The sum-based condition implies the simpler per-voxel condition c \textminus{} t > E + t, which the arithmetic check applies, so the certificates are conservative.

\begin{table}[tbp]
\caption{Overlap certificates. All organ constraints concern the mandible. $O$ denotes overlap, $m$ the cold-target allowance, and $n$ the hot-organ allowance. The dose margin is $c-L-0.10$ Gy.}\label{tab:S3}
\small
\setlength{\tabcolsep}{3.0pt}
\begin{tabular*}{\linewidth}{@{\extracolsep{\fill}}p{31.44pt}p{14.24pt}p{29.08pt}p{15.29pt}p{15.29pt}p{6.29pt}p{30.14pt}p{30.14pt}@{}}
\hline\hline
\noalign{\vskip 2pt}
\multicolumn{1}{@{}c}{\parbox[b]{31.44pt}{\raggedright Case\strut}} & \multicolumn{1}{c}{\parbox[b]{14.24pt}{\centering Set\strut}} & \multicolumn{1}{c}{\parbox[b]{29.08pt}{\centering Target\strut}} & \multicolumn{1}{c}{\parbox[b]{15.29pt}{\centering O\strut}} & \multicolumn{1}{c}{\parbox[b]{15.29pt}{\centering m\strut}} & \multicolumn{1}{c}{\parbox[b]{6.29pt}{\centering n\strut}} & \multicolumn{1}{c}{\parbox[b]{30.14pt}{\centering Voxel margin\strut}} & \multicolumn{1}{c@{}}{\parbox[b]{30.14pt}{\centering Dose margin\strut}} \tabularnewline
\hline
\noalign{\vskip 1pt}
\raggedright pt\_258 & \centering D & \centering 7000 & \centering 128 & \centering 48 & \centering 3 & \centering 77 & \centering 6.70 \tabularnewline
\raggedright pt\_306 & \centering D & \centering 7000 & \centering 104 & \centering 46 & \centering 3 & \centering 55 & \centering 6.70 \tabularnewline
\raggedright pt\_307 & \centering D & \centering 7000 & \centering 111 & \centering 49 & \centering 3 & \centering 59 & \centering 6.70 \tabularnewline
\raggedright pt\_320 & \centering D & \centering 7000 & \centering 127 & \centering 80 & \centering 3 & \centering 44 & \centering 6.70 \tabularnewline
\raggedright pt\_339 & \centering D & \centering 7000 & \centering 247 & \centering 64 & \centering 3 & \centering 180 & \centering 6.70 \tabularnewline
\noalign{\vskip 3pt}
\raggedright pt\_248 & \centering R & \centering 5600 & \centering 631 & \centering 286 & \centering 3 & \centering 342 & \centering 5.40 \tabularnewline
\raggedright pt\_252 & \centering R & \centering 7000 & \centering 231 & \centering 152 & \centering 5 & \centering 74 & \centering 6.70 \tabularnewline
\raggedright pt\_256 & \centering R & \centering 5600 & \centering 197 & \centering 108 & \centering 3 & \centering 86 & \centering 5.40 \tabularnewline
\raggedright pt\_264 & \centering R & \centering 7000 & \centering 280 & \centering 78 & \centering 4 & \centering 198 & \centering 6.70 \tabularnewline
\raggedright pt\_283 & \centering R & \centering 6300 & \centering 173 & \centering 151 & \centering 2 & \centering 20 & \centering 6.05 \tabularnewline
\raggedright pt\_299 & \centering R & \centering 7000 & \centering 66 & \centering 35 & \centering 3 & \centering 28 & \centering 6.70 \tabularnewline
\raggedright pt\_305 & \centering R & \centering 5600 & \centering 118 & \centering 89 & \centering 3 & \centering 26 & \centering 5.40 \tabularnewline
\raggedright pt\_312 & \centering R & \centering 7000 & \centering 74 & \centering 68 & \centering 4 & \centering 2 & \centering 6.70 \tabularnewline
\raggedright pt\_313 & \centering R & \centering 7000 & \centering 637 & \centering 83 & \centering 4 & \centering 550 & \centering 6.70 \tabularnewline
\raggedright pt\_334 & \centering R & \centering 7000 & \centering 330 & \centering 57 & \centering 3 & \centering 270 & \centering 6.70 \tabularnewline
\hline\hline
\end{tabular*}
\end{table}

\subsection{Feasible controls and independent dose reconstruction}\label{sec:S5.4}\nobreak

All 25 reserved feasible witnesses were reconstructed independently from the saved nonnegative fluence, and every hard goal was met with the same 0.05~Gy tolerance (\hyperref[tab:S4]{Table S4}). Three patients lacked SpinalCord, and two had no verified witness after the three-solve construction search; a failure to find a witness was treated as an eligibility exclusion and not as evidence of infeasibility.

\begin{table}[tbp]
\caption{Stored feasible witnesses. The minimum margin is the smallest signed distance to a tolerated hard-goal boundary. A positive value indicates that every hard goal is met.}\label{tab:S4}
\small
\setlength{\tabcolsep}{3.0pt}
\begin{tabular*}{\linewidth}{@{\extracolsep{\fill}}p{31.44pt}p{45.81pt}p{96.91pt}@{}}
\hline\hline
\noalign{\vskip 2pt}
\multicolumn{1}{@{}c}{\parbox[b]{31.44pt}{\raggedright Case\strut}} & \multicolumn{1}{c}{\parbox[b]{45.81pt}{\centering Hard goals\strut}} & \multicolumn{1}{c@{}}{\parbox[b]{96.91pt}{\centering Minimum margin (Gy)\strut}} \tabularnewline
\hline
\noalign{\vskip 1pt}
\raggedright pt\_242 & \centering 5 & \centering 2.5102 \tabularnewline
\raggedright pt\_245 & \centering 6 & \centering 0.5938 \tabularnewline
\raggedright pt\_248 & \centering 6 & \centering 2.3075 \tabularnewline
\raggedright pt\_249 & \centering 5 & \centering 2.5491 \tabularnewline
\raggedright pt\_251 & \centering 7 & \centering 3.5477 \tabularnewline
\raggedright pt\_256 & \centering 6 & \centering 2.9016 \tabularnewline
\raggedright pt\_263 & \centering 6 & \centering 2.9591 \tabularnewline
\raggedright pt\_264 & \centering 7 & \centering 2.2572 \tabularnewline
\raggedright pt\_267 & \centering 5 & \centering 0.0499 \tabularnewline
\raggedright pt\_275 & \centering 6 & \centering 3.1260 \tabularnewline
\raggedright pt\_278 & \centering 6 & \centering 2.1227 \tabularnewline
\raggedright pt\_282 & \centering 5 & \centering 3.5232 \tabularnewline
\raggedright pt\_287 & \centering 6 & \centering 2.5478 \tabularnewline
\raggedright pt\_292 & \centering 5 & \centering 1.2317 \tabularnewline
\raggedright pt\_299 & \centering 6 & \centering 3.3745 \tabularnewline
\raggedright pt\_303 & \centering 6 & \centering 2.8496 \tabularnewline
\raggedright pt\_304 & \centering 6 & \centering 0.9191 \tabularnewline
\raggedright pt\_305 & \centering 7 & \centering 3.5101 \tabularnewline
\raggedright pt\_312 & \centering 6 & \centering 3.4349 \tabularnewline
\raggedright pt\_317 & \centering 7 & \centering 3.3487 \tabularnewline
\raggedright pt\_319 & \centering 5 & \centering 0.4938 \tabularnewline
\raggedright pt\_326 & \centering 5 & \centering 3.0142 \tabularnewline
\raggedright pt\_328 & \centering 5 & \centering 1.4226 \tabularnewline
\raggedright pt\_334 & \centering 6 & \centering 0.4372 \tabularnewline
\raggedright pt\_337 & \centering 4 & \centering 1.9907 \tabularnewline
\hline\hline
\end{tabular*}
\end{table}

Independent reconstruction also checked all 327 submitted reserved routine and feasible-control plans using explicit order-statistic interpolation. The maximum discrepancy from the stored metrics was $4.41\times {10}^{-13}$ Gy, and every hard-goal fraction agreed. The witnesses demonstrate feasibility within the supplied fluence model; machine deliverability was not assessed.

\section{Detailed dosimetry, numerical verification, and system components}\label{sec:S6}\nobreak

\subsection{Submitted-plan dosimetry and attempt-level outcomes}\label{sec:S6.1}\nobreak

\begin{table*}[tp]
\caption{Reserved routine planning. Patient-mean dose endpoints among submitted plans, reported as mean (patients).}\label{tab:S5}
\small
\setlength{\tabcolsep}{3.0pt}
\begin{tabular*}{\linewidth}{@{\extracolsep{\fill}}p{79.23pt}p{43.40pt}p{46.29pt}p{46.33pt}p{43.76pt}@{}}
\hline\hline
\noalign{\vskip 2pt}
\multicolumn{1}{@{}c}{\parbox[b]{79.23pt}{\raggedright Endpoint (Gy)\strut}} & \multicolumn{1}{c}{\parbox[b]{43.40pt}{\centering Controller\strut}} & \multicolumn{1}{c}{\parbox[b]{46.29pt}{\centering Arithmetic\strut}} & \multicolumn{1}{c}{\parbox[b]{46.33pt}{\centering Interpreter\strut}} & \multicolumn{1}{c@{}}{\parbox[b]{43.76pt}{\centering Full agent\strut}} \tabularnewline
\hline
\noalign{\vskip 1pt}
\raggedright Mandible D0.1cc & \centering 69.33 \mbox{(19)} & \centering 69.83 \mbox{(19)} & \centering 69.43 \mbox{(20)} & \centering 69.30 \mbox{(20)} \tabularnewline
\raggedright Brainstem D0.1cc & \centering 43.90 \mbox{(25)} & \centering 43.89 \mbox{(25)} & \centering 44.13 \mbox{(26)} & \centering 40.01 \mbox{(26)} \tabularnewline
\raggedright Esophagus Dmean & \centering 39.24 \mbox{(13)} & \centering 38.56 \mbox{(14)} & \centering 38.56 \mbox{(14)} & \centering 32.40 \mbox{(14)} \tabularnewline
\raggedright External Dmax & \centering 75.89 \mbox{(29)} & \centering 75.90 \mbox{(29)} & \centering 75.92 \mbox{(30)} & \centering 76.76 \mbox{(30)} \tabularnewline
\raggedright Larynx Dmean & \centering 42.88 \mbox{(16)} & \centering 43.00 \mbox{(17)} & \centering 43.00 \mbox{(17)} & \centering 40.97 \mbox{(17)} \tabularnewline
\raggedright PTV56 D99 & \centering 54.92 \mbox{(26)} & \centering 54.97 \mbox{(26)} & \centering 54.92 \mbox{(27)} & \centering 55.00 \mbox{(27)} \tabularnewline
\raggedright PTV63 D99 & \centering 62.10 \mbox{(12)} & \centering 62.10 \mbox{(12)} & \centering 62.10 \mbox{(12)} & \centering 63.59 \mbox{(12)} \tabularnewline
\raggedright PTV70 D99 & \centering 66.68 \mbox{(29)} & \centering 66.47 \mbox{(29)} & \centering 66.56 \mbox{(30)} & \centering 67.89 \mbox{(30)} \tabularnewline
\raggedright Parotid\_L Dmean & \centering 25.53 \mbox{(29)} & \centering 25.54 \mbox{(29)} & \centering 25.53 \mbox{(30)} & \centering 25.41 \mbox{(30)} \tabularnewline
\raggedright Parotid\_R Dmean & \centering 25.60 \mbox{(29)} & \centering 25.59 \mbox{(29)} & \centering 25.61 \mbox{(30)} & \centering 25.32 \mbox{(30)} \tabularnewline
\raggedright SpinalCord D0.1cc & \centering 43.86 \mbox{(26)} & \centering 43.86 \mbox{(26)} & \centering 43.85 \mbox{(27)} & \centering 43.15 \mbox{(27)} \tabularnewline
\hline\hline
\end{tabular*}
\end{table*}

\begin{table*}[tp]
\caption{Reserved feasible controls. Patient-mean dose endpoints among submitted plans, reported as mean (patients).}\label{tab:S6}
\small
\setlength{\tabcolsep}{3.0pt}
\begin{tabular*}{\linewidth}{@{\extracolsep{\fill}}p{79.23pt}p{43.40pt}p{46.29pt}p{46.33pt}p{43.76pt}@{}}
\hline\hline
\noalign{\vskip 2pt}
\multicolumn{1}{@{}c}{\parbox[b]{79.23pt}{\raggedright Endpoint (Gy)\strut}} & \multicolumn{1}{c}{\parbox[b]{43.40pt}{\centering Controller\strut}} & \multicolumn{1}{c}{\parbox[b]{46.29pt}{\centering Arithmetic\strut}} & \multicolumn{1}{c}{\parbox[b]{46.33pt}{\centering Interpreter\strut}} & \multicolumn{1}{c@{}}{\parbox[b]{43.76pt}{\centering Full agent\strut}} \tabularnewline
\hline
\noalign{\vskip 1pt}
\raggedright Mandible D0.1cc & \centering 68.60 \mbox{(15)} & \centering 68.60 \mbox{(15)} & \centering 68.60 \mbox{(15)} & \centering 68.38 \mbox{(15)} \tabularnewline
\raggedright Brainstem D0.1cc & \centering 41.79 \mbox{(23)} & \centering 41.79 \mbox{(23)} & \centering 41.79 \mbox{(23)} & \centering 38.03 \mbox{(23)} \tabularnewline
\raggedright Esophagus Dmean & \centering 36.71 \mbox{(10)} & \centering 36.71 \mbox{(10)} & \centering 36.71 \mbox{(10)} & \centering 31.61 \mbox{(10)} \tabularnewline
\raggedright External Dmax & \centering 75.93 \mbox{(25)} & \centering 75.93 \mbox{(25)} & \centering 75.93 \mbox{(25)} & \centering 76.74 \mbox{(25)} \tabularnewline
\raggedright Larynx Dmean & \centering 42.34 \mbox{(13)} & \centering 42.34 \mbox{(13)} & \centering 42.34 \mbox{(13)} & \centering 39.45 \mbox{(13)} \tabularnewline
\raggedright PTV56 D99 & \centering 54.78 \mbox{(22)} & \centering 54.78 \mbox{(22)} & \centering 54.78 \mbox{(22)} & \centering 54.99 \mbox{(22)} \tabularnewline
\raggedright PTV63 D99 & \centering 62.92 \mbox{(9)} & \centering 62.92 \mbox{(9)} & \centering 62.92 \mbox{(9)} & \centering 63.23 \mbox{(9)} \tabularnewline
\raggedright PTV70 D99 & \centering 66.48 \mbox{(25)} & \centering 66.48 \mbox{(25)} & \centering 66.48 \mbox{(25)} & \centering 67.81 \mbox{(25)} \tabularnewline
\raggedright Parotid\_L Dmean & \centering 25.50 \mbox{(25)} & \centering 25.50 \mbox{(25)} & \centering 25.50 \mbox{(25)} & \centering 25.06 \mbox{(25)} \tabularnewline
\raggedright Parotid\_R Dmean & \centering 25.56 \mbox{(25)} & \centering 25.56 \mbox{(25)} & \centering 25.56 \mbox{(25)} & \centering 25.00 \mbox{(25)} \tabularnewline
\raggedright SpinalCord D0.1cc & \centering 34.59 \mbox{(25)} & \centering 34.59 \mbox{(25)} & \centering 34.59 \mbox{(25)} & \centering 34.14 \mbox{(25)} \tabularnewline
\hline\hline
\end{tabular*}
\end{table*}

\begin{table*}[tp]
\caption{Attempt-level dose accounting. All-hard-goal and rejection counts concern submitted plans, and the eligible denominators also include attempts without a submission.}\label{tab:S7}
\small
\setlength{\tabcolsep}{3.0pt}
\begin{tabular*}{\linewidth}{@{\extracolsep{\fill}}p{104.52pt}p{99.86pt}p{32.38pt}p{44.41pt}p{59.59pt}p{36.87pt}@{}}
\hline\hline
\noalign{\vskip 2pt}
\multicolumn{1}{@{}c}{\parbox[b]{104.52pt}{\raggedright Population\strut}} & \multicolumn{1}{c}{\parbox[b]{99.86pt}{\centering System\strut}} & \multicolumn{1}{c}{\parbox[b]{32.38pt}{\centering Eligible\strut}} & \multicolumn{1}{c}{\parbox[b]{44.41pt}{\centering Submitted\strut}} & \multicolumn{1}{c}{\parbox[b]{59.59pt}{\centering All hard goals\strut}} & \multicolumn{1}{c@{}}{\parbox[b]{36.87pt}{\centering Rejected\strut}} \tabularnewline
\hline
\noalign{\vskip 1pt}
\raggedright Reserved routine & \centering Controller & \centering 30 & \centering 29 & \centering 21 & \centering 0 \tabularnewline
\raggedright Reserved routine & \centering Interpreter + controller & \centering 60 & \centering 60 & \centering 42 & \centering 0 \tabularnewline
\raggedright Reserved routine & \centering Full agent & \centering 60 & \centering 59 & \centering 47 & \centering 0 \tabularnewline
\raggedright Reserved routine & \centering Arithmetic + controller & \centering 30 & \centering 29 & \centering 20 & \centering 0 \tabularnewline
\noalign{\vskip 3pt}
\raggedright Reserved feasible control & \centering Controller & \centering 25 & \centering 25 & \centering 19 & \centering 0 \tabularnewline
\raggedright Reserved feasible control & \centering Interpreter + controller & \centering 50 & \centering 50 & \centering 38 & \centering 0 \tabularnewline
\raggedright Reserved feasible control & \centering Full agent & \centering 50 & \centering 50 & \centering 38 & \centering 0 \tabularnewline
\raggedright Reserved feasible control & \centering Arithmetic + controller & \centering 25 & \centering 25 & \centering 19 & \centering 0 \tabularnewline
\noalign{\vskip 3pt}
\raggedright Development T1 & \centering Controller & \centering 50 & \centering 50 & \centering 25 & \centering 0 \tabularnewline
\raggedright Development T1 & \centering Heuristic & \centering 50 & \centering 50 & \centering 25 & \centering 0 \tabularnewline
\raggedright Development T1 & \centering Sonnet & \centering 50 & \centering 49 & \centering 20 & \centering 4 \tabularnewline
\raggedright Development T1 & \centering Full agent & \centering 50 & \centering 50 & \centering 22 & \centering 0 \tabularnewline
\raggedright Development T1 & \centering GLM & \centering 50 & \centering 50 & \centering 26 & \centering 3 \tabularnewline
\raggedright Development T1 & \centering Search & \centering 50 & \centering 50 & \centering 18 & \centering 2 \tabularnewline
\raggedright Development T1 & \centering Random & \centering 50 & \centering 50 & \centering 7 & \centering 19 \tabularnewline
\noalign{\vskip 3pt}
\raggedright Development T2 & \centering Controller & \centering 50 & \centering 50 & \centering 35 & \centering 0 \tabularnewline
\raggedright Development T2 & \centering Heuristic & \centering 50 & \centering 50 & \centering 30 & \centering 0 \tabularnewline
\raggedright Development T2 & \centering Sonnet & \centering 50 & \centering 50 & \centering 29 & \centering 5 \tabularnewline
\raggedright Development T2 & \centering Full agent & \centering 50 & \centering 50 & \centering 37 & \centering 0 \tabularnewline
\raggedright Development T2 & \centering GLM & \centering 50 & \centering 50 & \centering 28 & \centering 16 \tabularnewline
\raggedright Development T2 & \centering Search & \centering 50 & \centering 50 & \centering 23 & \centering 1 \tabularnewline
\raggedright Development T2 & \centering Random & \centering 50 & \centering 50 & \centering 13 & \centering 9 \tabularnewline
\hline\hline
\end{tabular*}
\end{table*}

\begin{table*}[tp]
\caption{The six technical failures and their membership in the primary endpoint.}\label{tab:S8}
\small
\setlength{\tabcolsep}{3.0pt}
\begin{tabular*}{\linewidth}{@{\extracolsep{\fill}}p{31.44pt}p{99.82pt}p{67.99pt}p{85.49pt}p{35.31pt}@{}}
\hline\hline
\noalign{\vskip 2pt}
\multicolumn{1}{@{}c}{\parbox[b]{31.44pt}{\raggedright Case\strut}} & \multicolumn{1}{c}{\parbox[b]{99.82pt}{\centering System\strut}} & \multicolumn{1}{c}{\parbox[b]{67.99pt}{\centering Scenario\strut}} & \multicolumn{1}{c}{\parbox[b]{85.49pt}{\centering Failure\strut}} & \multicolumn{1}{c@{}}{\parbox[b]{35.31pt}{\centering Primary\strut}} \tabularnewline
\hline
\noalign{\vskip 1pt}
\raggedright pt\_304 & \centering Controller & \centering Routine & \centering Interrupted attempt & \centering No \tabularnewline
\raggedright pt\_248 & \centering Arithmetic + controller & \centering Routine & \centering Interrupted attempt & \centering No \tabularnewline
\raggedright pt\_245 & \centering Arithmetic + controller & \centering Conflicting note & \centering Interrupted attempt & \centering No \tabularnewline
\raggedright pt\_251 & \centering Full agent & \centering Routine & \centering Connection failure & \centering No \tabularnewline
\raggedright pt\_267 & \centering Full agent & \centering Conflicting note & \centering Connection failure & \centering No \tabularnewline
\raggedright pt\_275 & \centering Full agent & \centering Coverage/cap & \centering Request timeout & \centering Yes \tabularnewline
\hline\hline
\end{tabular*}
\end{table*}

Tables \hyperref[tab:S5]{S5} and \hyperref[tab:S6]{S6} report conditional means over the patients with a submitted plan and the relevant structure, with patient counts in parentheses. Repetitions are averaged within each patient before patients receive equal weight. Because the available patient sets differ for some endpoints, these summaries are descriptive. \hyperref[tab:S7]{Table S7} reports the eligible attempt denominators, submissions, all-hard-goal attainment, and rejections, and no absent or failed submission is assigned a dose value.

Every system submitted every feasible control; however, all hard goals were met in only 19/25 attempts for each deterministic system and 38/50 for each language system, or 76\% for all systems. No submitted control crossed a PlanScore rejection gate. The conditional patient-mean PTV70 D99 was 66.48~Gy for the controller-based systems and 67.81~Gy for the full agent, and the corresponding spinal-cord D0.1cc values were 34.59 and 34.14~Gy, respectively. These differences in dose coexist with identical rates of submission and all-hard-goal attainment.

On the routine tasks, all hard goals were met in 21/30 controller, 20/30 arithmetic/controller, 42/60 interpreter/controller, and 47/60 full-agent attempts, and three routine attempts ended without a submission. These all-attempt denominators include nonsubmissions, whereas Tables \hyperref[tab:S5]{S5} and \hyperref[tab:S6]{S6} describe only submitted plans.

\subsection{Objective functions}\label{sec:S6.2}\nobreak

For structure dose vector $d$, priority $p$, nominal dose level $l$, and $N$ voxels including outside-mask zeros, the \texttt{min\_dose}, \texttt{max\_dose}, \texttt{mean\_dose}, and \texttt{uniform\_dose} penalties are, respectively,
\begin{gather*}
\frac{p}{N}\sum\limits_{i} {[l-{d}_{i}]}_{+}^{2}, \qquad \frac{p}{N}\sum\limits_{i} {[{d}_{i}-l]}_{+}^{2}, \\
p{[\bar{d}-l]}_{+}^{2}, \qquad \frac{p}{N}\sum\limits_{i} {({d}_{i}-l)}^{2},
\end{gather*}
where ${[x]}_{+}=\max(x,0)$. Priorities range from 0 to 1000, and these terms are penalties; they do not enforce hard constraints.

The optional dvh\_max term selects voxels above $l$ but at or below the dose quantile corresponding to the permitted hot-volume fraction, and it sums their squared excess with weight $p/N$, recomputing the selected set at each evaluation. This piecewise term is not treated as a globally convex dose-volume constraint. Smoothness adds $\lambda$ times the sum of squared differences between adjacent beamlets within each beam, with a default of $\lambda ={10}^{-4}$. Projection enforces nonnegative fluence, and a beamlet-weight cap is optional. Normalization after the solve rescales the fluence and can change every dose goal and complexity measure.

The default optimizer uses the fast iterative shrinkage-thresholding algorithm (FISTA) with backtracking and adaptive restart, and it stops at 500 iterations or at a relative tolerance of ${10}^{-4}$. This iteration limit defines the computational budget of the study. The verification below evaluates the implementation of the objective and the agreement of solvers on specified reduced problems, which is distinct from convergence at the full problem size.

\subsection{Reconstruction and optimizer verification}\label{sec:S6.3}\nobreak

Reconstruction from the published fluence on all 10 development cases agreed with the supplied dose to within $4.27\times {10}^{-14}$ Gy. Fifty synthetic checks compared the objective values and gradients of all five agent-selectable terms, outside-mask zeros, and smoothness with an independently implemented dense objective and with finite differences away from selection boundaries, and all met the specified tolerances of ${10}^{-9}$ for values and analytic gradients and ${10}^{-6}$ for finite differences. A diagonal least-squares problem with an analytic nonnegative solution had a maximum fluence error of $8.56\times {10}^{-8}$.

The agreement of the optimizer with an independently implemented L-BFGS-B solver was then checked on reduced versions of the 10 development cases, which sampled up to 60 voxels per structure (sampling seed 0) and retained the 32 beamlet columns with the largest norms, selected for each case independently of the solver outcomes. Both solvers minimized the flat-priority objective template of the controller, including the External term, and a comparison passed when the reference solver reported success,
\[
\frac{|{f}_{\mathrm{FISTA}}-{f}_{\mathrm{ref}}|}{\max(1,|{f}_{\mathrm{ref}}|)}\leq {10}^{-5},
\]
and both solutions had a projected-gradient residual no greater than ${10}^{-3}$, where the residual is the infinity norm of $w-\max(0,w-\nabla f)$.

All 10 comparisons passed, with a maximum scaled objective difference of $2.57\times {10}^{-13}$, a maximum projected residual of $1.07\times {10}^{-5}$, and a maximum voxel-dose difference of 0.000198~Gy. Two earlier series, which retained 128 evenly spaced columns and so left some directions weakly constrained, met the same criteria in 2 and 6 of the 10 cases, so the agreement on reduced problems depends on their conditioning. \hyperref[sec:S9.2]{Section S9.2} examines the effect of a longer optimization budget on the full-size problems.

\subsection{System components and available information}\label{sec:S6.4}\nobreak

The three controller-based systems share the planning controller, objective template, normalization candidates, and dose feedback. The base controller has no missing-structure preflight of its own; it escalates with the reason other after an objective naming an absent structure is rejected, and all 30 reserved missing-structure trajectories followed this path. These responses named the missing structure and received full credit in the decision accuracy, although their generic reason label would receive zero under the development T4 grading.

The arithmetic system checks for missing structures and the two contradiction families before planning. Its extra two-voxel safety margin and its use of rounded summary inputs differ from the exact task certificate, and it missed 1 of the 10 reserved overlap tasks. The interpreter uses one model call to escalate or delegate, with any flags appended to the submission of the controller, and the full agent chooses its own planning actions. Consequently, the comparisons evaluate complete systems that differ in their prompts, opportunities for action, and placement of information.

The case summary includes the case and track identifiers, structures, goals and their notes, prescriptions, physician notes, beam geometry, budgets, and disclosed rejection rules, and the full summary also contains voxel counts, rounded overlap fractions, and distances. Neither summary contains the transformation labels, goal-list name, construction log, certificate, expected decision, witness fluence, reference dose, or score weights. The full agent can request calculated dose-volume histograms and metrics, whereas the score-visible search is an explicitly privileged development comparator.

\section{Audit of the SAGE planning logs}\label{sec:S7}\nobreak

\subsection{Cohort and source records}\label{sec:S7.1}\nobreak

The single-institution records contained 37 initial planning runs from 31 patients and two refinement runs, which were started after an earlier run had failed to meet all of the loaded goals, and all 39 runs were audited. Each plan had a single target prescribed 21~Gy in one fraction. Every initial run began from a common objective template with every priority set to 1, and SAGE stopped a run once every loaded goal was met or after 10 numbered attempts. Of the 31 patients, 24 contributed 1 run, 6 contributed 2 runs, and 1 contributed 3 runs. None of the 39 starting states met all five goals. The 35 runs from 29 patients that met every loaded goal required one new dose calculation in 28 cases, two in 5, three in 1, and four in 1, and the other 4 runs reached the attempt limit. Each run had a reasoning and execution log and a dose-score workbook, and plans from the same patient shared a pseudonym. The audit describes these runs and is not an estimate of the success rate of the agent.

SAGE used QwQ-32B, run locally with Ollama (version 0.40.0; model tag qwq:32b with Q4\_K\_M quantization; context length of 32,768 tokens; temperature of 0.6; top-k of 40; no fixed random seed) on a server with four NVIDIA A100 graphics processing units. The treatment planning system (TPS) was Eclipse version 16.1, in which the plans were optimized with the Photon Optimizer and dose was calculated with the Analytical Anisotropic Algorithm (version 16.1.2) on a 1.25-mm grid; the logs recorded Photon Optimizer version 15.6.05 for 30 runs and version 16.1.2 for 9 runs.

Each of the four runs that reached the 10-attempt limit had one goal that failed at every new dose calculation (\hyperref[fig:4]{Figure 4} of the main manuscript): the Brain-GTV V12Gy (11.56--12.27~cm\textsuperscript{3} against <10~cm\textsuperscript{3}) in the initial run of Patient 27, the PTV V21Gy (74.70--82.35\% against >95\%) in its refinement run, and the PTV Dmax in both runs of Patient 30 (27.49--28.24~Gy and 26.90--28.11~Gy against <26.5~Gy). At the last measured state, these values were 11.74~cm\textsuperscript{3}, 81.59\%, 28.03~Gy, and 27.08~Gy, respectively. None of the four logs recorded a stopping reason, an escalation, or a manual action, and the instructions offered none. In the initial run of Patient 27, the reply in attempt 9 abbreviated its list of seven objectives with an ellipsis; the reply could not be read, and after a retry, only the first objective of the abbreviated list was transferred. The optimizer then reported that no lower objective had been defined, no dose was calculated in attempts 9 and 10, and the goal values of attempt 8 were reported again in both.

\subsection{Recorded goals, replies, flags, and execution}\label{sec:S7.2}\nobreak

All 39 runs loaded five goals: PTV V21Gy >95\%, PTV Dmax <26.5~Gy, brainstem Dmax <12.5~Gy, optic-chiasm Dmax <8~Gy, and Brain-GTV V12Gy <10~cm\textsuperscript{3}. Comparisons used the literal strict operators and the unrounded stored values, and a measured state met all goals only when every loaded goal passed.

The logs were parsed into the supplied objectives, the goal values shown to the model, the reply, the objective list read from it, the execution messages, and the subsequent goal measurements. For each recorded transfer, the structure, operator, dose, volume, and priority of every objective were compared between the final proposal and the list transferred to Eclipse. Each reply ends with a rationale whose first line states the failing goals; the goals named in this line, excluding any stated to pass, were compared with the goals that failed in the state shown to the model, judged with the strict operators from the printed values. The replies were also searched for the two flags defined in the instructions: ``no control,'' for a failing goal without a controlling objective, which appeared in the instructions of all 39 runs, and ``not feasible with priorities,'' for a goal that still failed after a stated number of attempts, which appeared in the instructions of the two refinement runs. A new dose calculation required a corresponding execution record, and a reply that could not be read did not add a measured dose state.

The 39 workbooks contained 610 goal-value rows, comprising 39 starting states and 83 subsequent measured states with five goals each. All values and goal flags matched the logs, and the recomputed threshold results agreed with the recorded flags. The 94 numbered attempts included 83 new dose calculations, 9 plan-copy failures without a dose calculation, which occurred across six goal-attaining runs, and the 2 attempts of Patient 27 in which the optimizer could not run. Of the three replies that could not be read, each followed by a new model query within the same attempt, two led to a successful objective transfer and a dose calculation, and one led to the single-objective transfer described in \hyperref[sec:S7.1]{Section S7.1}. Of the 85 recorded objective transfers, 84 matched the corresponding final proposals on the displayed fields; the exception was that single-objective transfer, whose proposal could not be read as a complete list.

In all 88 replies, the failing goals named in the rationale equaled the goals that failed in the state shown to the model. The ``no control'' flag was written once, in attempt 10 of the initial run of Patient 27, when the single remaining objective could not control the V12Gy goal. The ``not feasible with priorities'' flag was never written. Its condition was met in the refinement run of Patient 30, whose case-specific instructions asked for the flag if the PTV Dmax still failed after four attempts, and the PTV Dmax failed at all 10 calculations of that run. No prompt in any run stated the attempt number, and each prompt showed at most the three most recent attempts, so the attempt count on which this rule depended was not given to the model; in attempt 7, the reasoning that preceded the reply discussed the rule and counted the three attempts shown.

\subsection{Example of iterative planning}\label{sec:S7.3}\nobreak

\begin{table*}[tp]
\caption{Dose metrics of Patient 4 at its four new dose calculations. Values are rounded for display; the goals were evaluated with strict operators on the unrounded values. The retry after an unreadable reply in attempt 3 preceded the dose calculation of that attempt and added no dose state.}\label{tab:S9}
\small
\setlength{\tabcolsep}{3.0pt}
\begin{tabular*}{\linewidth}{@{\extracolsep{\fill}}p{36.74pt}p{48.71pt}p{49.10pt}p{49.10pt}p{56.11pt}p{56.37pt}@{}}
\hline\hline
\noalign{\vskip 2pt}
\multicolumn{1}{@{}c}{\parbox[b]{36.74pt}{\centering Attempt\strut}} & \multicolumn{1}{c}{\parbox[b]{48.71pt}{\centering PTV V21Gy (\%)\strut}} & \multicolumn{1}{c}{\parbox[b]{49.10pt}{\centering PTV Dmax (Gy)\strut}} & \multicolumn{1}{c}{\parbox[b]{49.10pt}{\centering Brainstem Dmax (Gy)\strut}} & \multicolumn{1}{c}{\parbox[b]{56.11pt}{\centering Optic-chiasm Dmax (Gy)\strut}} & \multicolumn{1}{c@{}}{\parbox[b]{56.37pt}{\centering Brain-GTV V12Gy (cm\textsuperscript{3})\strut}} \tabularnewline
\hline
\noalign{\vskip 1pt}
\centering 1 & \centering 21.75 & \centering 22.18 & \centering 0.81 & \centering 0.049 & \centering 3.25 \tabularnewline
\centering 2 & \centering 60.71 & \centering 23.03 & \centering 1.82 & \centering 0.058 & \centering 3.43 \tabularnewline
\centering 3 & \centering 87.97 & \centering 23.61 & \centering 1.59 & \centering 0.061 & \centering 3.32 \tabularnewline
\centering 4 & \centering 96.62 & \centering 25.66 & \centering 0.91 & \centering 0.057 & \centering 3.79 \tabularnewline
\hline\hline
\end{tabular*}
\ogtablenote{Loaded thresholds: PTV V21Gy >95\%; PTV Dmax <26.5 Gy; brainstem Dmax <12.5 Gy; optic-chiasm Dmax <8 Gy; Brain-GTV V12Gy <10 cm\textsuperscript{3}. Coverage was the only unmet goal at calculations 1--3; all five goals were met at calculation 4.}
\end{table*}

Patient 4 was selected because its trajectory contained multiple measured states and a retry after an unreadable reply. \hyperref[fig:4]{Figure 4(a)} of the main manuscript shows its target coverage at each new calculation, and \hyperref[tab:S9]{Table S9} lists all five loaded dose metrics at each calculation.

Over four new calculations, PTV V21Gy rose from 21.75\% to 96.62\% while the other four loaded goals passed at every calculation, and the log recorded that planning stopped once all five goals were met.

Across the four updates that produced a calculation, the priority of the target lower-dose objective remained 300, that of the upper-dose objective decreased from 100 to 70, 50, and 35, and those of the three ring objectives decreased from 50 at the first update to 18--25 at the fourth, while all seven organ-objective priorities remained 20. The target objectives (21 and 25~Gy) and the three ring objectives (1.6~Gy at 58.07\% volume, 4.2~Gy at 19.38\%, and 10.5~Gy at 0\%) kept their dose and volume parameters across these updates, so the dose changes followed the priority updates, although the contribution of any single change was not isolated.

Within attempt 3, the log recorded a reply that could not be read, a retry with a new model query, a complete objective list, the transferred objectives, and a new dose calculation, in that order. The unreadable reply produced no new dose state and was not counted as an additional attempt, and all four effective proposals in this case matched the transferred and workbook objectives.

\section{Further analyses of the reserved evaluation}\label{sec:S8}\nobreak

The analyses in this section use the stored outcomes of all 1,080 scheduled entries, the stored certificates and witnesses, and, for submission notes and interpreter explanations only, the retained trajectories. No model was called and no plan was recomputed. Patient-bootstrap intervals used 10,000 resamples.

\subsection{Task families, weighting, and the contrast between the language systems}\label{sec:S8.1}\nobreak

\begin{table*}[tp]
\caption{Decision accuracy (\%) by task family and under three weightings. Family values are patient means with 95\% patient-bootstrap intervals where any attempt failed. The primary aggregate averages the eligible families within each patient; equal family weights average the four family rates; equal task weights average the 94 patient-family tasks. The final row gives differences in percentage points, with the interval of the primary analysis for the primary weighting.}\label{tab:S10}
\small
\setlength{\tabcolsep}{3.0pt}
\begin{tabular*}{\linewidth}{@{\extracolsep{\fill}}p{114.24pt}p{39.59pt}p{63.36pt}p{48.91pt}p{35.72pt}p{47.98pt}p{70.22pt}p{47.98pt}@{}}
\hline\hline
\noalign{\vskip 2pt}
\multicolumn{1}{@{}c}{\parbox[b]{114.24pt}{\raggedright System\strut}} & \multicolumn{1}{c}{\parbox[b]{39.59pt}{\centering Missing structure (30)\strut}} & \multicolumn{1}{c}{\parbox[b]{63.36pt}{\centering Coverage/cap (29)\strut}} & \multicolumn{1}{c}{\parbox[b]{48.91pt}{\centering Overlap (10)\strut}} & \multicolumn{1}{c}{\parbox[b]{35.72pt}{\centering Feasible control (25)\strut}} & \multicolumn{1}{c}{\parbox[b]{47.98pt}{\centering Primary\strut}} & \multicolumn{1}{c}{\parbox[b]{70.22pt}{\centering Equal family\strut}} & \multicolumn{1}{c@{}}{\parbox[b]{47.98pt}{\centering Equal task\strut}} \tabularnewline
\hline
\noalign{\vskip 1pt}
\raggedright Controller & \centering 100 & \centering 0 & \centering 0 & \centering 100 & \centering 60.0 & \centering 50.0 & \centering 58.5 \tabularnewline
\raggedright Arithmetic checks + controller & \centering 100 & \centering 100 & \centering 90 \mbox{(70--100)} & \centering 100 & \centering 99.2 & \centering 97.5 & \centering 98.9 \tabularnewline
\raggedright Interpreter + controller & \centering 100 & \centering 100 & \centering 50 \mbox{(25--75)} & \centering 100 & \centering 95.4 & \centering 87.5 & \centering 94.7 \tabularnewline
\raggedright Full LLM agent & \centering 100 & \centering 96.6 \mbox{(91.4--100)} & \centering 25 \mbox{(10--40)} & \centering 100 & \centering 92.1 & \centering 80.4 & \centering 91.0 \tabularnewline
\raggedright Interpreter minus full agent & \centering 0 & \centering 3.4 & \centering 25 & \centering 0 & \centering 3.33 \mbox{(0.14--7.08)} & \centering 7.11 \mbox{(\textminus{}1.08 to 15.69)} & \centering 3.72 \mbox{(0.00--7.61)} \tabularnewline
\hline\hline
\end{tabular*}
\end{table*}

\begin{figure*}[tp]
\centering
\includegraphics[width=178mm]{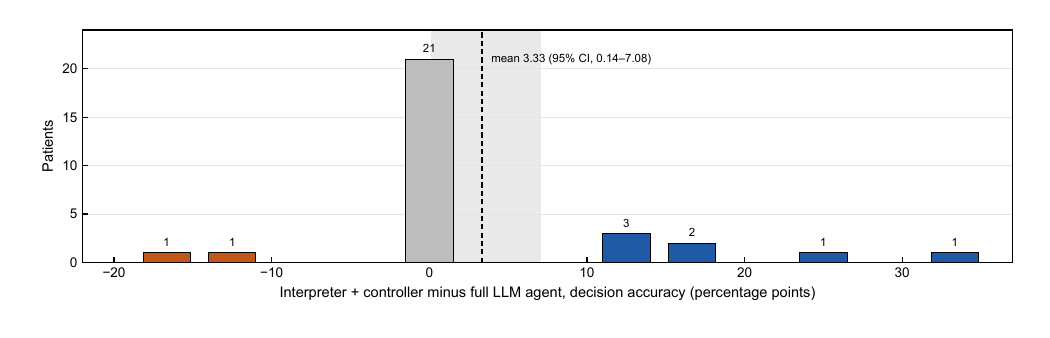}
\caption{Distribution of the patient-level differences in decision accuracy between the interpreter/controller and the full LLM agent, after repetitions and eligible families were averaged: 7 patients favored the interpreter/controller, 21 were tied, and 2 favored the full agent. The dashed line marks the mean difference of 3.33 percentage points and the shaded band its 95\% confidence interval (0.14--7.08).}\label{fig:S5}
\end{figure*}

\hyperref[tab:S10]{Table S10} gives the decision accuracy by task family with 95\% patient-bootstrap intervals, together with the aggregate under the primary weighting and under equal weights for the four families or for the 94 patient-family tasks. Under the primary weighting, the missing-structure and feasible-control families, which every system passed, carried 34.2\% and 25.8\% of each aggregate, the coverage/cap family 30.8\%, and the overlap family 9.2\%. Of the 3.33-point interpreter-minus-agent difference (\hyperref[fig:S5]{Figure S5}), 2.22 points came from the overlap family, 0.56 from one coverage/cap task on which one run of the full agent submitted a plan (pt\_249), and 0.56 from the full-agent timeout (pt\_275). An exact paired sign-flip test of the 30 patient differences gave a two-sided p value of 0.105. With the timed-out attempt treated as missing rather than as a failure, the difference was 2.78 percentage points, with a 95\% interval of \textminus{}0.28 to 6.39 (p = 0.17).

\subsection{Overlap contradictions task by task}\label{sec:S8.2}\nobreak

\begin{table*}[tp]
\caption{Reserved overlap contradictions by task. O, m, and n are the overlap, cold-target, and hot-organ voxel counts of the certificate (\hyperref[tab:S3]{Table S3}), and the count margin is O \textminus{} (m + n). Runs 1 and 2 are the two repetitions of each language system.}\label{tab:S11}
\small
\setlength{\tabcolsep}{3.0pt}
\begin{tabular*}{\linewidth}{@{\extracolsep{\fill}}p{31.84pt}p{15.29pt}p{15.29pt}p{6.29pt}p{30.14pt}p{46.29pt}p{46.33pt}p{46.33pt}p{44.41pt}p{44.41pt}@{}}
\hline\hline
\noalign{\vskip 2pt}
\multicolumn{1}{@{}c}{\parbox[b]{31.84pt}{\raggedright Patient\strut}} & \multicolumn{1}{c}{\parbox[b]{15.29pt}{\centering O\strut}} & \multicolumn{1}{c}{\parbox[b]{15.29pt}{\centering m\strut}} & \multicolumn{1}{c}{\parbox[b]{6.29pt}{\centering n\strut}} & \multicolumn{1}{c}{\parbox[b]{30.14pt}{\centering Count margin\strut}} & \multicolumn{1}{c}{\parbox[b]{46.29pt}{\centering Arithmetic\strut}} & \multicolumn{1}{c}{\parbox[b]{46.33pt}{\centering Interpreter run 1\strut}} & \multicolumn{1}{c}{\parbox[b]{46.33pt}{\centering Interpreter run 2\strut}} & \multicolumn{1}{c}{\parbox[b]{44.41pt}{\centering Full agent run 1\strut}} & \multicolumn{1}{c@{}}{\parbox[b]{44.41pt}{\centering Full agent run 2\strut}} \tabularnewline
\hline
\noalign{\vskip 1pt}
\raggedright pt\_248 & \centering 631 & \centering 286 & \centering 3 & \centering 342 & \centering Escalated & \centering Escalated & \centering Escalated & \centering Escalated & \centering Submitted \tabularnewline
\raggedright pt\_252 & \centering 231 & \centering 152 & \centering 5 & \centering 74 & \centering Escalated & \centering Submitted & \centering Submitted & \centering Escalated & \centering Submitted \tabularnewline
\raggedright pt\_256 & \centering 197 & \centering 108 & \centering 3 & \centering 86 & \centering Escalated & \centering Submitted & \centering Submitted & \centering Submitted & \centering Submitted \tabularnewline
\raggedright pt\_264 & \centering 280 & \centering 78 & \centering 4 & \centering 198 & \centering Escalated & \centering Escalated & \centering Escalated & \centering Submitted & \centering Submitted \tabularnewline
\raggedright pt\_283 & \centering 173 & \centering 151 & \centering 2 & \centering 20 & \centering Escalated & \centering Escalated & \centering Escalated & \centering Submitted & \centering Submitted \tabularnewline
\raggedright pt\_299 & \centering 66 & \centering 35 & \centering 3 & \centering 28 & \centering Escalated & \centering Escalated & \centering Submitted & \centering Submitted & \centering Submitted \tabularnewline
\raggedright pt\_305 & \centering 118 & \centering 89 & \centering 3 & \centering 26 & \centering Escalated & \centering Submitted & \centering Escalated & \centering Submitted & \centering Submitted \tabularnewline
\raggedright pt\_312 & \centering 74 & \centering 68 & \centering 4 & \centering 2 & \centering Submitted & \centering Submitted & \centering Submitted & \centering Escalated & \centering Submitted \tabularnewline
\raggedright pt\_313 & \centering 637 & \centering 83 & \centering 4 & \centering 550 & \centering Escalated & \centering Escalated & \centering Submitted & \centering Submitted & \centering Escalated \tabularnewline
\raggedright pt\_334 & \centering 330 & \centering 57 & \centering 3 & \centering 270 & \centering Escalated & \centering Submitted & \centering Escalated & \centering Escalated & \centering Submitted \tabularnewline
\hline\hline
\end{tabular*}
\end{table*}

\hyperref[tab:S11]{Table S11} lists each reserved overlap task with its certificate and the final decision of every attempt. Working from the rounded summary fields, the arithmetic check escalated nine tasks and missed pt\_312, whose count margin was 2 voxels. The interpreter delegated both runs of pt\_252, pt\_256, and pt\_312 and split its runs on pt\_299, pt\_305, pt\_313, and pt\_334; each time it delegated an overlap task, its explanation stated that no incompatibility among the hard requirements had been demonstrated. The full agent escalated five attempts, never both runs of the same task. Each of its 15 overlap submissions followed all three optimizations and listed the unmet hard goals, and none described the request as infeasible. The two runs agreed on 6 of 10 overlap tasks for the interpreter and 5 of 10 for the full agent, and on every task of the other primary families except two coverage/cap tasks of the full agent, one of which contained the timeout.

\subsection{Final decisions and the conflicting-note scenario}\label{sec:S8.3}\nobreak

\begin{table*}[tp]
\caption{Final decision of every eligible reserved attempt. Escalation reasons are the labels recorded by each system. No attempt ended in an automatic submission or a forced ending.}\label{tab:S12}
\small
\setlength{\tabcolsep}{3.0pt}
\begin{tabular*}{\linewidth}{@{\extracolsep{\fill}}p{132.04pt}p{75.73pt}p{42.75pt}p{73.87pt}p{46.67pt}p{43.50pt}p{59.44pt}@{}}
\hline\hline
\noalign{\vskip 2pt}
\multicolumn{1}{@{}c}{\parbox[b]{132.04pt}{\raggedright System\strut}} & \multicolumn{1}{c}{\parbox[b]{75.73pt}{\centering Scenario\strut}} & \multicolumn{1}{c}{\parbox[b]{42.75pt}{\centering Attempts\strut}} & \multicolumn{1}{c}{\parbox[b]{73.87pt}{\centering Escalated (reason)\strut}} & \multicolumn{1}{c}{\parbox[b]{46.67pt}{\centering Submitted\strut}} & \multicolumn{1}{c}{\parbox[b]{43.50pt}{\centering Technical failure\strut}} & \multicolumn{1}{c@{}}{\parbox[b]{59.44pt}{\centering Median optimizations\strut}} \tabularnewline
\hline
\noalign{\vskip 1pt}
\raggedright Controller & \centering Routine & \centering 30 & \centering 0 & \centering 29 & \centering 1 & \centering 2.0 \tabularnewline
\raggedright Controller & \centering Missing structure & \centering 30 & \centering 30 (other) & \centering 0 & \centering 0 & \centering 0.0 \tabularnewline
\raggedright Controller & \centering Conflicting note & \centering 30 & \centering 0 & \centering 30 & \centering 0 & \centering 2.0 \tabularnewline
\raggedright Controller & \centering Coverage/cap & \centering 29 & \centering 0 & \centering 29 & \centering 0 & \centering 3.0 \tabularnewline
\raggedright Controller & \centering Overlap & \centering 10 & \centering 0 & \centering 10 & \centering 0 & \centering 3.0 \tabularnewline
\raggedright Controller & \centering Feasible control & \centering 25 & \centering 0 & \centering 25 & \centering 0 & \centering 2.0 \tabularnewline
\noalign{\vskip 3pt}
\raggedright Arithmetic checks + controller & \centering Routine & \centering 30 & \centering 0 & \centering 29 & \centering 1 & \centering 2.0 \tabularnewline
\raggedright Arithmetic checks + controller & \centering Missing structure & \centering 30 & \centering 30 (missing structure) & \centering 0 & \centering 0 & \centering 0.0 \tabularnewline
\raggedright Arithmetic checks + controller & \centering Conflicting note & \centering 30 & \centering 0 & \centering 29 & \centering 1 & \centering 2.0 \tabularnewline
\raggedright Arithmetic checks + controller & \centering Coverage/cap & \centering 29 & \centering 29 (infeasible) & \centering 0 & \centering 0 & \centering 0.0 \tabularnewline
\raggedright Arithmetic checks + controller & \centering Overlap & \centering 10 & \centering 9 (infeasible) & \centering 1 & \centering 0 & \centering 0.0 \tabularnewline
\raggedright Arithmetic checks + controller & \centering Feasible control & \centering 25 & \centering 0 & \centering 25 & \centering 0 & \centering 2.0 \tabularnewline
\noalign{\vskip 3pt}
\raggedright Interpreter + controller & \centering Routine & \centering 60 & \centering 0 & \centering 60 & \centering 0 & \centering 2.0 \tabularnewline
\raggedright Interpreter + controller & \centering Missing structure & \centering 60 & \centering 60 (missing structure) & \centering 0 & \centering 0 & \centering 0.0 \tabularnewline
\raggedright Interpreter + controller & \centering Conflicting note & \centering 60 & \centering 31 (contradictory instructions) & \centering 29 & \centering 0 & \centering 0.0 \tabularnewline
\raggedright Interpreter + controller & \centering Coverage/cap & \centering 58 & \centering 58 (infeasible) & \centering 0 & \centering 0 & \centering 0.0 \tabularnewline
\raggedright Interpreter + controller & \centering Overlap & \centering 20 & \centering 10 (infeasible) & \centering 10 & \centering 0 & \centering 1.5 \tabularnewline
\raggedright Interpreter + controller & \centering Feasible control & \centering 50 & \centering 0 & \centering 50 & \centering 0 & \centering 2.0 \tabularnewline
\noalign{\vskip 3pt}
\raggedright Full LLM agent & \centering Routine & \centering 60 & \centering 0 & \centering 59 & \centering 1 & \centering 2.0 \tabularnewline
\raggedright Full LLM agent & \centering Missing structure & \centering 60 & \centering 60 (missing structure) & \centering 0 & \centering 0 & \centering 0.0 \tabularnewline
\raggedright Full LLM agent & \centering Conflicting note & \centering 60 & \centering 0 & \centering 59 & \centering 1 & \centering 2.0 \tabularnewline
\raggedright Full LLM agent & \centering Coverage/cap & \centering 58 & \centering 52 (infeasible); 4 (contradictory instructions) & \centering 1 & \centering 1 & \centering 0.0 \tabularnewline
\raggedright Full LLM agent & \centering Overlap & \centering 20 & \centering 5 (infeasible) & \centering 15 & \centering 0 & \centering 3.0 \tabularnewline
\raggedright Full LLM agent & \centering Feasible control & \centering 50 & \centering 0 & \centering 50 & \centering 0 & \centering 1.5 \tabularnewline
\hline\hline
\end{tabular*}
\end{table*}

\hyperref[tab:S12]{Table S12} gives the final decision of every eligible attempt. No attempt ended in an automatic submission or a forced ending. Every escalation by the arithmetic system and by both language systems carried a reason label accepted for its task family, whereas the controller escalated every missing-structure task with the generic label other after its objective was rejected. In the conflicting-note scenario, the automatic grader credited an escalation or a submission that flagged the note. The interpreter/controller was credited on all 60 attempts (31 escalations and 29 flagged submissions), the full agent on 59 of 60 (all flagged submissions, with one technical failure), and neither scripted system on any attempt, since neither responds to the notes.

\subsection{Feasible controls with unmet hard goals}\label{sec:S8.4}\nobreak

\begin{table*}[tp]
\caption{Unmet hard goals of submitted feasible controls. The controller-based systems submitted the same plan for each control. Beyond tolerance is the shortfall or excess outside the 0.05~Gy goal tolerance; the witness margin is the stored witness's distance inside the tolerated boundary of the same goal.}\label{tab:S13}
\small
\setlength{\tabcolsep}{3.0pt}
\begin{tabular*}{\linewidth}{@{\extracolsep{\fill}}p{31.84pt}p{69.92pt}p{18.59pt}p{67.70pt}p{39.12pt}p{22.71pt}p{61.72pt}p{53.24pt}@{}}
\hline\hline
\noalign{\vskip 2pt}
\multicolumn{1}{@{}c}{\parbox[b]{31.84pt}{\raggedright Patient\strut}} & \multicolumn{1}{c}{\parbox[b]{69.92pt}{\centering System\strut}} & \multicolumn{1}{c}{\parbox[b]{18.59pt}{\centering Run\strut}} & \multicolumn{1}{c}{\parbox[b]{67.70pt}{\centering Unmet hard goal\strut}} & \multicolumn{1}{c}{\parbox[b]{39.12pt}{\centering Achieved (Gy)\strut}} & \multicolumn{1}{c}{\parbox[b]{22.71pt}{\centering Goal (Gy)\strut}} & \multicolumn{1}{c}{\parbox[b]{61.72pt}{\centering Beyond tolerance (Gy)\strut}} & \multicolumn{1}{c@{}}{\parbox[b]{53.24pt}{\centering Witness margin (Gy)\strut}} \tabularnewline
\hline
\noalign{\vskip 1pt}
\raggedright pt\_245 & \centering Controller-based & \centering All & \centering External Dmax & \centering 78.63 & \centering 77.00 & \centering 1.58 & \centering 0.59 \tabularnewline
\raggedright pt\_245 & \centering Controller-based & \centering All & \centering PTV\_7000 D99 & \centering 58.36 & \centering 66.50 & \centering 8.09 & \centering 2.94 \tabularnewline
\raggedright pt\_245 & \centering Controller-based & \centering All & \centering PTV\_6300 D99 & \centering 58.10 & \centering 59.85 & \centering 1.70 & \centering 6.99 \tabularnewline
\raggedright pt\_278 & \centering Controller-based & \centering All & \centering PTV\_7000 D99 & \centering 63.45 & \centering 66.50 & \centering 3.00 & \centering 3.31 \tabularnewline
\raggedright pt\_287 & \centering Controller-based & \centering All & \centering PTV\_7000 D99 & \centering 64.44 & \centering 66.50 & \centering 2.01 & \centering 3.39 \tabularnewline
\raggedright pt\_292 & \centering Controller-based & \centering All & \centering PTV\_7000 D99 & \centering 64.38 & \centering 66.50 & \centering 2.07 & \centering 3.44 \tabularnewline
\raggedright pt\_304 & \centering Controller-based & \centering All & \centering PTV\_7000 D99 & \centering 62.83 & \centering 66.50 & \centering 3.62 & \centering 3.02 \tabularnewline
\raggedright pt\_334 & \centering Controller-based & \centering All & \centering PTV\_7000 D99 & \centering 64.65 & \centering 66.50 & \centering 1.80 & \centering 3.17 \tabularnewline
\raggedright pt\_334 & \centering Controller-based & \centering All & \centering PTV\_5600 D99 & \centering 50.68 & \centering 53.20 & \centering 2.47 & \centering 2.85 \tabularnewline
\noalign{\vskip 3pt}
\raggedright pt\_245 & \centering Full LLM agent & \centering 1 & \centering PTV\_7000 D99 & \centering 59.84 & \centering 66.50 & \centering 6.61 & \centering 2.94 \tabularnewline
\raggedright pt\_245 & \centering Full LLM agent & \centering 1 & \centering PTV\_6300 D99 & \centering 58.89 & \centering 59.85 & \centering 0.91 & \centering 6.99 \tabularnewline
\raggedright pt\_245 & \centering Full LLM agent & \centering 1 & \centering PTV\_5600 D99 & \centering 51.24 & \centering 53.20 & \centering 1.91 & \centering 5.56 \tabularnewline
\raggedright pt\_245 & \centering Full LLM agent & \centering 2 & \centering PTV\_7000 D99 & \centering 56.75 & \centering 66.50 & \centering 9.70 & \centering 2.94 \tabularnewline
\raggedright pt\_245 & \centering Full LLM agent & \centering 2 & \centering PTV\_6300 D99 & \centering 54.44 & \centering 59.85 & \centering 5.36 & \centering 6.99 \tabularnewline
\raggedright pt\_245 & \centering Full LLM agent & \centering 2 & \centering PTV\_5600 D99 & \centering 47.89 & \centering 53.20 & \centering 5.26 & \centering 5.56 \tabularnewline
\raggedright pt\_278 & \centering Full LLM agent & \centering 1 & \centering External Dmax & \centering 80.50 & \centering 77.00 & \centering 3.45 & \centering 2.12 \tabularnewline
\raggedright pt\_278 & \centering Full LLM agent & \centering 1 & \centering PTV\_7000 D99 & \centering 64.11 & \centering 66.50 & \centering 2.34 & \centering 3.31 \tabularnewline
\raggedright pt\_292 & \centering Full LLM agent & \centering 1 & \centering PTV\_7000 D99 & \centering 65.18 & \centering 66.50 & \centering 1.27 & \centering 3.44 \tabularnewline
\raggedright pt\_304 & \centering Full LLM agent & \centering 1 & \centering External Dmax & \centering 79.02 & \centering 77.00 & \centering 1.97 & \centering 0.92 \tabularnewline
\raggedright pt\_304 & \centering Full LLM agent & \centering 1 & \centering PTV\_7000 D99 & \centering 66.25 & \centering 66.50 & \centering 0.20 & \centering 3.02 \tabularnewline
\raggedright pt\_304 & \centering Full LLM agent & \centering 2 & \centering External Dmax & \centering 80.16 & \centering 77.00 & \centering 3.11 & \centering 0.92 \tabularnewline
\raggedright pt\_319 & \centering Full LLM agent & \centering 1 & \centering PTV\_7000 D99 & \centering 64.33 & \centering 66.50 & \centering 2.12 & \centering 3.21 \tabularnewline
\raggedright pt\_319 & \centering Full LLM agent & \centering 2 & \centering External Dmax & \centering 80.23 & \centering 77.00 & \centering 3.18 & \centering 0.49 \tabularnewline
\raggedright pt\_328 & \centering Full LLM agent & \centering 1 & \centering External Dmax & \centering 80.05 & \centering 77.00 & \centering 3.00 & \centering 1.42 \tabularnewline
\raggedright pt\_328 & \centering Full LLM agent & \centering 2 & \centering External Dmax & \centering 79.31 & \centering 77.00 & \centering 2.26 & \centering 1.42 \tabularnewline
\raggedright pt\_334 & \centering Full LLM agent & \centering 1 & \centering External Dmax & \centering 79.15 & \centering 77.00 & \centering 2.10 & \centering 0.44 \tabularnewline
\raggedright pt\_334 & \centering Full LLM agent & \centering 1 & \centering PTV\_7000 D99 & \centering 65.85 & \centering 66.50 & \centering 0.60 & \centering 3.17 \tabularnewline
\raggedright pt\_334 & \centering Full LLM agent & \centering 1 & \centering PTV\_5600 D99 & \centering 52.43 & \centering 53.20 & \centering 0.72 & \centering 2.85 \tabularnewline
\raggedright pt\_334 & \centering Full LLM agent & \centering 2 & \centering External Dmax & \centering 78.37 & \centering 77.00 & \centering 1.32 & \centering 0.44 \tabularnewline
\hline\hline
\end{tabular*}
\end{table*}

\hyperref[tab:S13]{Table S13} lists every hard goal that a submitted feasible control left unmet. The controller, the arithmetic system, and both runs of the interpreter submitted the same plan for each control, so their 6, 6, and 12 deficient attempts concern the same six patients; the full agent's 12 deficient attempts concern 7 patients. The unmet goals were target coverage (the D99 of PTV70 and, in two patients, also that of PTV63 or PTV56) and the External maximum dose. Every deficient submission listed its unmet hard goals in its submission note. The stored witnesses met the same goals with margins of 2.9 to 3.4~Gy for the PTV70 D99 and 0.4 to 2.1~Gy for the External maximum dose. \hyperref[sec:S9.2]{Section S9.2} tests whether a longer optimization budget closes these gaps.

\subsection{Alternative action policies}\label{sec:S8.5}\nobreak

\begin{table*}[tp]
\caption{Patient-weighted decision accuracy (\%) under the primary and two alternative policies, with the interpreter-minus-agent difference in percentage points and its 95\% patient-bootstrap interval.}\label{tab:S14}
\small
\setlength{\tabcolsep}{3.0pt}
\begin{tabular*}{\linewidth}{@{\extracolsep{\fill}}p{225.50pt}p{45.83pt}p{48.73pt}p{48.76pt}p{26.20pt}p{84.98pt}@{}}
\hline\hline
\noalign{\vskip 2pt}
\multicolumn{1}{@{}c}{\parbox[b]{225.50pt}{\raggedright Policy\strut}} & \multicolumn{1}{c}{\parbox[b]{45.83pt}{\centering Controller\strut}} & \multicolumn{1}{c}{\parbox[b]{48.73pt}{\centering Arithmetic checks + controller\strut}} & \multicolumn{1}{c}{\parbox[b]{48.76pt}{\centering Interpreter + controller\strut}} & \multicolumn{1}{c}{\parbox[b]{26.20pt}{\centering Full LLM agent\strut}} & \multicolumn{1}{c@{}}{\parbox[b]{84.98pt}{\centering Interpreter minus full agent\strut}} \tabularnewline
\hline
\noalign{\vskip 1pt}
\raggedright Primary policy & \centering 60.0 & \centering 99.2 & \centering 95.4 & \centering 92.1 & \centering 3.33 \mbox{(0.14--7.08)} \tabularnewline
\raggedright Escalation reason must match the family & \centering 25.8 & \centering 99.2 & \centering 95.4 & \centering 92.1 & \centering 3.33 \mbox{(0.14--7.08)} \tabularnewline
\raggedright Feasible control credited only if every hard goal is met & \centering 53.6 & \centering 92.8 & \centering 89.0 & \centering 85.7 & \centering 3.33 \mbox{(\textminus{}2.08 to 8.75)} \tabularnewline
\hline\hline
\end{tabular*}
\end{table*}

\begin{table*}[tp]
\caption{Specification of the planning environment. Values are those of the reserved evaluation.}\label{tab:S15}
\small
\setlength{\tabcolsep}{3.0pt}
\begin{tabular*}{\linewidth}{@{\extracolsep{\fill}}p{82.87pt}p{421.13pt}@{}}
\hline\hline
\noalign{\vskip 2pt}
\multicolumn{1}{@{}c}{\parbox[b]{82.87pt}{\raggedright Item\strut}} & \multicolumn{1}{c@{}}{\parbox[b]{421.13pt}{\raggedright Specification\strut}} \tabularnewline
\hline
\noalign{\vskip 1pt}
\raggedright Beams and grid & \raggedright Nine coplanar 6-MV beams at 40\textdegree{} spacing, as described for OpenKBP-Opt; 5~\texttimes{}~5~mm beamlets; voxels of 2.73--5.12~mm in plane and 1.5--3.27~mm between slices \tabularnewline
\raggedright Hard goals & \raggedright External Dmax \ensuremath{\le}77~Gy; PTV70, PTV63, and PTV56 D99 \ensuremath{\ge}66.5, 59.85, and 53.2~Gy; brainstem, spinal cord, and mandible D0.1cc \ensuremath{\le}50, 45, and 73.5~Gy \tabularnewline
\raggedright Soft goals & \raggedright Parotid Dmean \ensuremath{\le}26~Gy; esophagus and larynx Dmean \ensuremath{\le}45~Gy \tabularnewline
\raggedright Tolerances and gates & \raggedright 0.05~Gy for dose-valued goals; rejection when the global maximum exceeds 115\% of the highest prescription or a target D99 falls below 80\% of its prescription, with a boundary tolerance of $10^{-8}$~Gy \tabularnewline
\raggedright Missing-structure task & \raggedright Adds a hard D0.1cc goal of 54~Gy on the left optic nerve, a structure that is absent \tabularnewline
\raggedright Feasible control & \raggedright Spinal-cord D0.1cc limit at the midpoint between 45~Gy and the witness cord dose, rounded to 0.1~Gy \mbox{(33.6--39.4~Gy in 24 reserved tasks; 45.0~Gy for pt\_267)} \tabularnewline
\raggedright Witness search & \raggedright FISTA, at most three 500-iteration solves; hard-goal terms at priority 1,000 and soft terms at 1; cord level lowered to 0.75 times the achieved cord dose between solves; no normalization \tabularnewline
\raggedright Controller objective & \raggedright For each target, a minimum dose at the prescription and a ceiling at 1.07 times it; maximum-dose terms at each D0.1cc and Dmax limit; mean-dose terms at each Dmean limit; all priorities 1 \tabularnewline
\raggedright Controller back-off & \raggedright Priorities of unmet terms multiplied by 3 for organs and 2 for targets, up to 1,000; unmet upper limits lowered in proportion to the excess; target ceilings lowered when the External maximum is exceeded \tabularnewline
\raggedright Normalization candidates & \raggedright The best plan as optimized; rescaled so that the highest target D99 equals its goal; rescaled so that the External Dmax equals 76.923~Gy when exceeded; chosen by rejection status, then hard goals met, then soft violation \tabularnewline
\raggedright PlanScore R & \raggedright R = \mbox{(1 + mean r)}/2, where r = clip((reference \textminus{} achieved)/reference, \textminus{}1, 1) over the soft mean-dose goals of non-target structures with a positive reference mean dose; R = 0.5 when none is eligible \tabularnewline
\raggedright Calibration on reference plans & \raggedright Scoring the 100 reference plans led to the External goal of 77~Gy and the 115\% gate; the escalation credit of 0.9 is the score of a plan meeting every goal at reference organ doses \tabularnewline
\raggedright Full summary & \raggedright For each structure, the voxel count, the volume to 0.01~cm\textsuperscript{3}, the overlap fraction with each target to four decimal places, and the minimum distance to each target to 0.1~mm; beam geometry; goals, notes, rules, and budget \tabularnewline
\raggedright Reduced summary & \raggedright Structure names and volumes, prescriptions, goals, notes, and rules \tabularnewline
\raggedright Arithmetic check & \raggedright Coverage/cap: escalate when c \textminus{} 0.05 > E + 0.05~Gy. Overlap: escalate when the overlap estimated from the rounded fraction exceeds the cold-target and hot-organ allowances by more than 2 voxels \tabularnewline
\hline\hline
\end{tabular*}
\end{table*}

\begin{table*}[tp]
\caption{Execution settings of the reserved evaluation with Terra, from its design and run records.}\label{tab:S16}
\small
\setlength{\tabcolsep}{3.0pt}
\begin{tabular*}{\linewidth}{@{\extracolsep{\fill}}p{114.58pt}p{389.42pt}@{}}
\hline\hline
\noalign{\vskip 2pt}
\multicolumn{1}{@{}c}{\parbox[b]{114.58pt}{\raggedright Item\strut}} & \multicolumn{1}{c@{}}{\parbox[b]{389.42pt}{\raggedright Setting\strut}} \tabularnewline
\hline
\noalign{\vskip 1pt}
\raggedright Gateway and endpoint & \raggedright OpenRouter, model route openai/gpt-5.6-terra, restricted to the OpenAI endpoint openai/gpt-5.6-terra-20260709; fallback providers disabled \tabularnewline
\raggedright Campaign & \raggedright 9--10 September 2026, in three segments separated by two interruptions; 1,441 model requests, of which 1,438 returned a response \tabularnewline
\raggedright Output limits & \raggedright 2,048 tokens for the interpreter and 8,192 per call for the full agent; the longest completion was 1,103 tokens \tabularnewline
\raggedright Sampling & \raggedright Provider seeds 0 and 1; temperature not sent, because the endpoint did not support it \tabularnewline
\raggedright Retries and caching & \raggedright No retries; response caching disabled; the campaign stopped on any cache hit or provider error other than a malformed completion \tabularnewline
\raggedright Timeout & \raggedright The 180-s default of the HTTP client \tabularnewline
\raggedright Replies not in the required format & \raggedright Interpreter: none of 308 replies. Full agent: none of 1,130 calls; one reply contained no planning action and was followed by an escalation \tabularnewline
\raggedright Freeze & \raggedright Checksum manifest of 37 files \mbox{(SHA-256 29d819a5\ldots{}f627)}, committed before the evaluation \tabularnewline
\hline\hline
\end{tabular*}
\end{table*}

\hyperref[tab:S14]{Table S14} recalculates the aggregate under two alternative policies. Requiring the escalation reason to match the task family lowered only the controller, from 60.0\% to 25.8\%, because its missing-structure escalations used the generic reason. Crediting a feasible-control submission only when every hard goal was met lowered every system by 6.4 percentage points, and the interpreter-minus-agent difference remained 3.33 percentage points, with a 95\% interval of \textminus{}2.08 to 8.75.

\section{Second language model, longer optimization budgets, and fluence complexity}\label{sec:S9}\nobreak

The experiments in this section were run on the reserved tasks. Credit, weighting, and bootstrap intervals follow the primary analysis, and no comparison among these runs was tested statistically.

\subsection{Rebuilt tasks and their verification}\label{sec:S9.1}\nobreak

For these experiments, the 40 development and reserved patients were imported again from the same OpenKBP-Opt archive. The 180 task inputs of the reserved patients and all 54 certificate counts were rebuilt exactly, and the 217 goal values of the stored feasible-control solutions and the 2,852 goal values of the submitted reserved plans, recomputed from the stored fluences, equaled the stored values. Every run checked its task inputs and prompt against the original tasks.

\subsection{Controller with longer optimization budgets}\label{sec:S9.2}\nobreak

\begin{table*}[tp]
\caption{Deficient feasible controls of the controller at three optimization budgets. The unmet hard goals were identical at all budgets, and the six controls are the same at each. The 500-iteration values are those of the main evaluation.}\label{tab:S17}
\small
\setlength{\tabcolsep}{3.0pt}
\begin{tabular*}{\linewidth}{@{\extracolsep{\fill}}p{31.84pt}p{210.12pt}p{80.83pt}p{63.36pt}p{63.36pt}@{}}
\hline\hline
\noalign{\vskip 2pt}
\multicolumn{1}{@{}c}{\parbox[b]{31.84pt}{\raggedright Patient\strut}} & \multicolumn{1}{c}{\parbox[b]{210.12pt}{\centering Unmet hard goals at every budget\strut}} & \multicolumn{1}{c}{\parbox[b]{80.83pt}{\centering PTV\_7000 D99, 500 iterations (Gy)\strut}} & \multicolumn{1}{c}{\parbox[b]{63.36pt}{\centering 2,000 iterations (Gy)\strut}} & \multicolumn{1}{c@{}}{\parbox[b]{63.36pt}{\centering 5,000 iterations (Gy)\strut}} \tabularnewline
\hline
\noalign{\vskip 1pt}
\raggedright pt\_245 & \centering PTV\_7000 D99, PTV\_6300 D99, External Dmax & \centering 58.36 & \centering 58.37 & \centering 58.37 \tabularnewline
\raggedright pt\_278 & \centering PTV\_7000 D99 & \centering 63.45 & \centering 63.46 & \centering 63.46 \tabularnewline
\raggedright pt\_287 & \centering PTV\_7000 D99 & \centering 64.44 & \centering 64.45 & \centering 64.45 \tabularnewline
\raggedright pt\_292 & \centering PTV\_7000 D99 & \centering 64.38 & \centering 64.38 & \centering 64.38 \tabularnewline
\raggedright pt\_304 & \centering PTV\_7000 D99 & \centering 62.83 & \centering 62.82 & \centering 62.82 \tabularnewline
\raggedright pt\_334 & \centering PTV\_7000 D99, PTV\_5600 D99 & \centering 64.65 & \centering 64.63 & \centering 64.63 \tabularnewline
\hline\hline
\end{tabular*}
\end{table*}

The controller was rerun on the 25 reserved feasible controls with 2,000 and 5,000 iterations per optimization, with its budget of three optimizations, objective template, adjustment rules, and normalization unchanged and with the relative stopping tolerance of $10^{-4}$ and patience of 10 iterations retained. With 500 iterations, 37 of the 58 optimizations had stopped at the iteration limit. With 2,000 iterations, 1 of 57 optimizations reached the limit, and with 5,000, none of 57 did; in both, the median optimization stopped at the tolerance after 444 iterations, and the longest used 2,355. Every control was submitted at every budget, and the same 19 controls met every hard goal. \hyperref[tab:S17]{Table S17} lists the six deficient controls, whose unmet hard goals were identical at all three budgets and whose PTV\_7000 D99 changed by at most 0.02~Gy. Since the stored witnesses had met the same goals with at most three 500-iteration solves, the objectives of the controller, and not the iteration budget, limited these plans.

\subsection{Ministral 3 8B in the two language systems}\label{sec:S9.3}\nobreak

\begin{table*}[tp]
\caption{Decision accuracy (\%) of the two language systems run with Ministral 3 8B on the 94 primary reserved tasks. Family values are patient means with 95\% patient-bootstrap intervals, and the patient mean is the primary aggregate. In the lower rows, an escalation counted only when its reason named the problem of its family (the first alternative policy of \hyperref[sec:S8.5]{Section S8.5}). Technical failures, including replies that did not follow the required format, received no credit.}\label{tab:S18}
\small
\setlength{\tabcolsep}{3.0pt}
\begin{tabular*}{\linewidth}{@{\extracolsep{\fill}}p{118.56pt}p{67.20pt}p{67.20pt}p{67.20pt}p{67.20pt}p{31.84pt}p{43.62pt}@{}}
\hline\hline
\noalign{\vskip 2pt}
\multicolumn{1}{@{}c}{\parbox[b]{118.56pt}{\raggedright System\strut}} & \multicolumn{1}{c}{\parbox[b]{67.20pt}{\centering Missing structure (30)\strut}} & \multicolumn{1}{c}{\parbox[b]{67.20pt}{\centering Coverage/cap (29)\strut}} & \multicolumn{1}{c}{\parbox[b]{67.20pt}{\centering Overlap (10)\strut}} & \multicolumn{1}{c}{\parbox[b]{67.20pt}{\centering Feasible control (25)\strut}} & \multicolumn{1}{c}{\parbox[b]{31.84pt}{\centering Patient mean\strut}} & \multicolumn{1}{c@{}}{\parbox[b]{43.62pt}{\centering Technical failures\strut}} \tabularnewline
\hline
\noalign{\vskip 1pt}
\raggedright Interpreter + controller & \centering 60.0 \mbox{(48.3--70.0)} & \centering 22.4 \mbox{(13.8--31.0)} & \centering 30.0 \mbox{(15.0--45.0)} & \centering 10.0 \mbox{(2.0--18.0)} & \centering 31.7 & \centering 113 of 188 \tabularnewline
\raggedright Full LLM agent & \centering 63.3 \mbox{(50.0--76.7)} & \centering 13.8 \mbox{(5.2--24.1)} & \centering 20.0 \mbox{(5.0--35.0)} & \centering 64.0 \mbox{(50.0--78.0)} & \centering 43.5 & \centering 28 of 188 \tabularnewline
\raggedright Reason required: interpreter & \centering 58.3 & \centering 10.3 & \centering 5.0 & \centering 10.0 &  &  \tabularnewline
\raggedright Reason required: full agent & \centering 55.0 & \centering 13.8 & \centering 20.0 & \centering 64.0 &  &  \tabularnewline
\hline\hline
\end{tabular*}
\end{table*}

Both language systems were run on the 94 primary tasks with Ministral 3 8B (Mistral AI), an openly available model with 8.9 billion parameters, run locally with Ollama (version 0.40.3; model tag ministral-3:8b with Q4\_K\_M quantization; context length of 16,384 tokens; default temperature of 0.15) on a desktop computer with an Apple M4 processor and 16 GB of memory. With seeds 0 and 1, each system made 188 attempts (\hyperref[tab:S18]{Table S18} and \hyperref[fig:2]{Figure 2} of the main manuscript). The request timeout was raised from 180 to 900 s for local inference, and a planning action that the local software could not read counted as a technical failure of the attempt. Because any interpreter reply that does not follow the required format counts as a technical failure, the decision stated in each interpreter reply was also read with the format requirement relaxed. The replies of the interpreter did not follow the required format in 113 attempts: 72 gave an empty escalation reason together with a decision to plan, 40 returned JSON that could not be read, and 1 was empty. Read with the format requirement relaxed, the replies decided to plan on 9 of the 20 overlap attempts and escalated 6, and none of the six explanations mentioned the mandible. Under the alternative policy that required a matching escalation reason, the decision accuracy of the interpreter was 58.3\%, 10.3\%, 5.0\%, and 10.0\% on the missing-structure, coverage/cap, overlap, and feasible-control families, respectively.

The full agent lost 26 attempts to planning actions that the local software could not read, and two further attempts were ended by the harness. The agent escalated 4 of the 20 overlap attempts, each after one to three optimizations whose plans had missed hard goals and without reference to the overlapping voxels, and it submitted plans on 12. It also submitted plans on 16 of the 60 missing-structure attempts and escalated 7 of the 50 feasible-control attempts, six of them as infeasible, and 9 of its 32 submitted feasible controls met every hard goal. Its median episode used 7 model calls (maximum, 11).

\subsection{Fluence complexity of the reserved submissions}\label{sec:S9.4}\nobreak

\begin{table*}[tp]
\caption{Fluence complexity of the submitted reserved plans. SPG is the sum of positive gradients as defined in OpenKBP-Opt, and the OpenKBP-Opt reference plans were optimized with SPG \ensuremath{\le} 65 and beamlet weights \ensuremath{\le} 15. The controller-based systems submitted identical feasible-control plans.}\label{tab:S19}
\small
\setlength{\tabcolsep}{3.0pt}
\begin{tabular*}{\linewidth}{@{\extracolsep{\fill}}p{129.78pt}p{23.94pt}p{84.71pt}p{38.74pt}p{146.06pt}p{49.93pt}@{}}
\hline\hline
\noalign{\vskip 2pt}
\multicolumn{1}{@{}c}{\parbox[b]{129.78pt}{\raggedright System\strut}} & \multicolumn{1}{c}{\parbox[b]{23.94pt}{\centering Plans\strut}} & \multicolumn{1}{c}{\parbox[b]{84.71pt}{\centering SPG, median (IQR)\strut}} & \multicolumn{1}{c}{\parbox[b]{38.74pt}{\centering SPG \ensuremath{\le} 65\strut}} & \multicolumn{1}{c}{\parbox[b]{146.06pt}{\centering Maximum beamlet weight, median\strut}} & \multicolumn{1}{c@{}}{\parbox[b]{49.93pt}{\centering Weight \ensuremath{\le} 15\strut}} \tabularnewline
\hline
\noalign{\vskip 1pt}
\multicolumn{6}{@{}l}{\textit{Routine and feasible-control plans}} \tabularnewline
\raggedright Controller & \centering 54 & \centering 136.5 \mbox{(115.6--162.9)} & \centering 0 & \centering 14.70 & \centering 27 \tabularnewline
\raggedright Arithmetic checks + controller & \centering 54 & \centering 137.1 \mbox{(121.1--163.4)} & \centering 0 & \centering 14.70 & \centering 27 \tabularnewline
\raggedright Interpreter + controller & \centering 110 & \centering 136.8 \mbox{(115.6--163.4)} & \centering 0 & \centering 15.06 & \centering 54 \tabularnewline
\raggedright Full LLM agent & \centering 109 & \centering 380.9 \mbox{(250.2--454.8)} & \centering 0 & \centering 26.45 & \centering 13 \tabularnewline
\noalign{\vskip 4pt}
\multicolumn{6}{@{}l}{\textit{All submitted plans}} \tabularnewline
\raggedright Controller & \centering 123 & \centering 141.2 \mbox{(123.7--181.2)} & \centering 0 & \centering 15.77 & \centering 55 \tabularnewline
\raggedright Arithmetic checks + controller & \centering 84 & \centering 136.5 \mbox{(114.0--163.2)} & \centering 0 & \centering 15.06 & \centering 41 \tabularnewline
\raggedright Interpreter + controller & \centering 149 & \centering 137.3 \mbox{(120.3--184.0)} & \centering 0 & \centering 15.77 & \centering 69 \tabularnewline
\raggedright Full LLM agent & \centering 184 & \centering 390.0 \mbox{(266.1--463.0)} & \centering 0 & \centering 27.37 & \centering 19 \tabularnewline
\hline\hline
\end{tabular*}
\end{table*}

The sum of positive gradients (SPG) and the maximum beamlet weight were computed for every submitted reserved plan with a stored fluence, with the definitions of the development experiments, which follow the OpenKBP-Opt code: for each beam and fluence row, the positive differences between adjacent beamlets were summed, the largest row sum of each beam was taken, and these values were summed over the beams. \hyperref[tab:S19]{Table S19} summarizes the 540 submitted plans and the 327 routine and feasible-control plans. No plan met the SPG limit of 65, and the smallest value was 85.3. The median SPG of the full agent was about 2.8 times that of the controller-based systems, and its plans exceeded the beamlet-weight cap of 15 more often. The values describe the normalized fluences that were submitted, since the fluences before normalization were not stored.

\end{document}